%% file: journal_arxiv.tex
\documentclass[11pt]{article}
\usepackage{fullpage}
\usepackage[utf8]{inputenc}
\usepackage[T1]{fontenc}
\usepackage[latin9]{luainputenc}
\usepackage[authoryear,round]{natbib}
 \usepackage[margin=1in]{geometry}
\usepackage{amsmath, amsthm, amssymb}
\usepackage{setspace}

\usepackage{macros}

\begin{document}

\title{Best of Many Worlds Guarantees for Online Learning with Knapsacks}

\author{Andrea Celli\thanks{Equal contribution.}\\
Bocconi University\\
\texttt{andrea.celli2@unibocconi.it}
\and
Matteo Castiglioni $^\ast$\\
Politecnico di Milano\\
\texttt{matteo.castiglioni@polimi.it}
\and 
Christian Kroer\\
Columbia University\\
\texttt{christian.kroer@columbia.edu}
}

\date{\today}
\maketitle

\begin{abstract}
    We study online learning problems in which a decision maker wants to maximize their expected reward without violating a finite set of $m$ resource constraints. By casting the learning process over a suitably defined space of \emph{strategy mixtures}, we recover strong duality on a Lagrangian relaxation of the underlying optimization problem, even for general settings with non-convex reward and resource-consumption functions.
    Then, we provide the first best-of-many-worlds type framework for this setting, with no-regret guarantees under stochastic, adversarial, and non-stationary inputs. 
    Our framework yields the same regret guarantees of prior work in the stochastic case. On the other hand, when budgets grow at least linearly in the time horizon, it allows us to provide a \emph{constant} competitive ratio in the adversarial case, which improves over the best known upper bound bound of $O(\log m\log T)$.
    Moreover, our framework allows the decision maker to handle non-convex reward and cost functions. We provide two game-theoretic applications of our framework to give further evidence of its flexibility. In doing so, we show that it can be employed to implement budget-pacing mechanisms in repeated first-price auctions.
\end{abstract}

\blfootnote{A preliminary version of this paper was published at the 2022 International Conference on Machine Learning as \citep{castiglioni2022online}.
The present version includes new sections on nonstationary inputs and budget management in first-price auctions, as well as updates to the presentation throughout the paper.}

\vspace{0.2cm}
\setstretch{1.5}

\input{intro}

\input{related_work}

\input{preliminaries}

\input{mixtures}

\input{alg}

\input{adversarial}

\input{stochastic}

\input{nonstationary}

\input{appl}

\input{pacing}

\setstretch{1}
\bibliographystyle{plainnat}
{\small \bibliography{bib-abbrv,refs}}

\input{tools}

\end{document}

%% file: intro.tex
\section{Introduction}

In this paper, we study online learning problems in which the decision maker has to satisfy supply or budget constraints. In particular, the decision maker is endowed with $m\ge 1$ limited resources which are consumed over time. For each round $t$ up to the time horizon $T$, the decision maker chooses a strategy $\vxi_t$ which defines a probability measure over the set of available actions. Then, they observe some feedback about the reward and resource consumption incurred by playing $\vxi_t$. 
The process stops at time horizon $T$, or when the total consumption of some resource exceeds its budget $B$. The goal is to maximize the total reward. Our framework can be instantiated both in the \emph{full-information feedback} as well as in the \emph{bandit feedback} setting.
In the full information feedback setting, the decision maker observes the reward function $f_t$ and resource-consumption function $c_t$ at each $t$. In the bandit feedback setting they only get to observe $(f_t(\vx_t),c_t(\vx_t))$, where $\vx_t$ is the realized action selected according to $\vxi_t$.

Our framework builds on the well-known \emph{Bandits with Knapsacks} (BwK) model originally introduced by \citet{badanidiyuru2013bandits,Badanidiyuru2018jacm}, which has various motivating applications such as dynamic pricing~\citep{besbes2009dynamic,babaioff2012dynamic,besbes2012blind,wang2014close}, dynamic procurement~\citep{badanidiyuru2012learning,singla2013truthful}, and online ad allocation~\citep{slivkins2013dynamic,combes2015bandits}. 
In all such settings, it is common for the decision maker to have resource constraint on, for example, the overall amount of money which can be spent in a certain time window.

The BwK framework was first studied in the stochastic setting, in which inputs $(f_t,c_t)$ are drawn from some underlying unknown distribution. In i.i.d. settings it is possible to prove sublinear regret guarantees with respect to the best dynamic policy~\citep{badanidiyuru2012learning,agrawal2014bandits}.
In subsequent work, \citet{immorlica2019adversarial} introduced the \emph{Adversarial Bandits with Knapsacks} framework, which extends the classic model of \emph{adversarial bandits}~\citep{auer2002nonstochastic} by considering the ``pessimistic'' scenario in which inputs can be arbitrary. 
The Adversarial BwK setting is inherently more difficult that its stochastic counterpart, since the decision maker has to decide how much budget to save for the future, without being able to predict it. 
In general, in Adversarial BwK regret minimization is no longer possible \citep{immorlica2019adversarial}. Therefore, the goal in such setting is to design algorithms with \emph{small competitive ratio}, which is defined as the ratio between the reward provided by the best fixed policy, and the (expected) reward of the algorithm.

In this work, we focus on the regime in which $B=\Omega(T)$, that is the budget grows at least linearly in the time horizon $T$. This is the case, for example, when the decision maker has a fixed per-iteration budget as in most motivating applications, such as online advertising (see, \emph{e.g.}, \citet{balseiro2022best}). 
In this setting, we resolve the following two open questions posed by~\citet{immorlica2019adversarial}:
\begin{OneLiners}
    \item \emph{Is it possible to solve both stochastic and adversarial version of BwK with exactly the same algorithm?}
    \item \emph{Is it possible to obtain a constant-factor competitive ratio in the adversarial case for the regime $B=\Omega(T)$?}
\end{OneLiners}
We answer both questions positively. 
Our analysis is based on a primal-dual learning algorithm, as it is customarily done in online problems with packing constraints. We cast the learning process over a suitably defined space of strategy mixtures, which allows us to recover strong duality of the Lagrangian relaxation of the underlying optimization problem even when $f_t$ and $c_t$ are non-convex functions, and the set of available actions $\cX$ may be non-convex and non-compact. This strictly generalizes the setting studied by \citet{immorlica2019adversarial}.

\xhdr{Best-of-many-worlds guarantees.} 
We provide the first \emph{best-of-many-worlds} algorithm for online learning problems with knapsacks. This result allows our framework to achieve good worst-case performance, while being able to take advantage of \emph{well-behaved} problem instances. This makes progress on the line of work initiated by \citet{bubeck2012best}. 
In particular, we show that our meta-algorithm (\cref{alg:meta alg}) guarantees a tight regret bound in the stochastic case (\cref{thm:stochastic}), matching known results by \citet{badanidiyuru2013bandits,agrawal2014bandits,immorlica2019adversarial}. Moreover, in the adversarial case, it guarantees a constant approximation ratio which is computed as the maximum single-round resource consumption over the per-iteration budget (\cref{thm:adversarial}). This improves over the $O(m\log T)$ ratio by \citet{immorlica2019adversarial}, and over the recent $O(\log m \log T)$ ratio by \citet{kesselheim2020online}. 
Finally, we study the non-stationary input model by \citet{balseiro2022best}, in which requests are drawn from independent but not necessarily identical distributions at each time $t$, and show that our meta-algorithm is robust to adversarial corruptions to i.i.d. inputs (\cref{thm:nonstat}). This model generalizes the stochastic input model.

\xhdr{Applications.} One of the advantages of our analysis is its ``modularity''. The meta-algorithms does not have any particular requirement on the primal and dual regret minimizers that have to be employed. Therefore, the framework can be applied to any setting for which there exists a suitable primal regret minimizer. In order to provide further evidence of the flexibility of our framework, we present a few applications. 
First, we provide an explicit instantiation of our meta-algorithm in the classical BwK setting by \citet{badanidiyuru2013bandits}. Then, we focus on two game-theoretic applications of particular interest. 
First, we show that our framework may be employed to extend the work by \citet{balcan2015commitment} to repeated Stackelberg Security Games in which resources are costly, and the planner must satisfy some resource-consumption constraints. 
This has applications, for example, to the protection of critical infrastructure sites~\citep{tambe2011security}.
Finally, we argue that our framework can be adapted to handle budget-pacing mechanisms~\citep{conitzer2021multiplicative,balseiro2021budget} in the context of repeated first-price auctions. This is particularly relevant for modern auction markets operated by large Internet advertising companies. First we present a simplified analysis in a setting in which valuations and bids are finite, and then we extend it to the general ``continuous settings''. 
We observe that previous work only addressed the case of budget pacing in  repeated second-price auctions \citep{balseiro2019learning}.

%% file: related_work.tex
\section{Related Work}

We survey the most relevant works with respect to ours. For further background on online learning and regret-minimizing algorithms the reader can refer to the monographs by \citet{cesa2006prediction,hazan2016introduction,slivkins2019introduction}.

\xhdr{Bandits with Knapsacks.} The Stochastic \emph{Bandits with Knapsacks} (BwK) framework was introduced and optimally solved by~\citet{badanidiyuru2013bandits,Badanidiyuru2018jacm}.
Other regret-optimal algorithms for Stochastic BwK have been proposed by \citet{agrawal2014bandits,agrawal2019bandits}, and by \citet{immorlica2019adversarial}.
The BwK framework has been subsequently extended to numerous settings such as, for example, more general notions of resources and constraints~\citep{agrawal2014bandits,agrawal2019bandits}, contextual bandits~\citep{dudik2011efficient,badanidiyuru2014resourceful,agarwal2014taming,agrawal2016efficient}, and combinatorial semi-bandits \citep{sankararaman2018combinatorial}.
Finally, various work have studied applications with budget/supply constraints which can be seen as special cases of the Stochastic BwK framework such as, for example, dynamic pricing \citep{besbes2009dynamic,babaioff2015dynamic,besbes2012blind,wang2014close}, dynamic procurement \citep{badanidiyuru2012learning}, dynamic ad allocation \citep{combes2015bandits, balseiro2019learning}.

\xhdr{Adversarial BwK.} 
The \emph{Adversarial Bandits with Knapsacks} setting was first studied by \citet{immorlica2019adversarial}, who proved a $O(m\log T)$ competitive ratio for the case in which the sequence of rewards and costs is chosen by an oblivious adversary.
To achieve such result, the authors introduce a new algorithm for the stochastic BwK setting, called \texttt{LagrangeBwK}, which provides a natural game-theoretic interpretation of Stochastic Bwk, while achieving optimal rate of convergence. Such algorithm is then used as a subroutine to achieve the $O(m\log T)$ competitive ratio in the adversarial setting, which can be shown to hold also in high probability. 
\texttt{LagrangeBwK} has an intuitive primal-dual interpretation since arms can be thought of as primal variables, and resources as dual variables. Then, \texttt{LagrangeBwK} basically sets up a repeated two-player zero-sum game between a primal and dual player. 
Our algorithm (see \cref{alg:meta alg}) is version of \texttt{LagrangeBwK} with the following key modifications: it is instantiated on the space of \emph{strategy mixtures} (see \cref{sec: mixtures}); we don't employ a \emph{time resource}  but we explicitly manage the stopping time within the algorithm; the dual regret minimizer directly outputs a dual vector in $[0,1/\rho]^m$ instead of sampling a single resource.
We highlight that \citet{immorlica2019adversarial} also show that no algorithm can achieve a competitive ratio better than $O(\log T)$ on all problem instances, using a lower-bounding construction with only two arms and a single resource. 
Recently, \citet{kesselheim2020online} refined the analysis for the general Adersarial BwK setting to obtain an $O(\log m\,\log T)$ competitive ratio. They also prove that such competitive ratio is optimal up to constant factors.
Finally, we observe that well-known online packing problems can be seen as special case of Adversarial BwK with a more permissive feedback model, which allows the decision maker to observe full feedback before choosing an action (see, e.g., \citet{buchbinder2009online,Buchbinder2009design,devanur2011near}). 
In online packing settings, since the decision maker is endowed with more information at the time of taking decisions, it is possible to derive $O(\log T)$ competitive ratio guarantees against the optimal dynamic policy. 
Online packing subsumes various online matching problems which are relevant for applications in settings such as ad allocation in online-advertising platforms \citep{mehta2007adwords}. In particular, \citet{mehta2007adwords} and \citet{buchbinder2007online} study the \emph{AdWords} problem (i.e., an online matching problem in which rewards are proportional to resource consumption) and provide an algorithm yielding a $(1-1/e)$-fraction of the optimal allocation in hindsight, which is optimal. 
\citet{feldman2009online} provide the same guarantees under the free disposal assumption. 
Finally, \citet{balseiro2020dual,balseiro2022best} consider general online allocation problems (i.e., without the free disposal assumption), and prove that it is possible to attain a (parametric) constant-factor competitive ratio under adversarial inputs. 
Our result for adversarial inputs (\cref{thm:adversarial}) shows that, when budgets grow at least linearly in time, it is possible to attain a constant-factor competitive ratio even in the Adversarial BwK setting, in which the feedback $(f_t,c_t)$ is observed only \emph{after} the action at time $t$ has been selected.

\xhdr{Best-of-both-worlds algorithms.} 
\emph{Best-of-both-worlds-type} algorithms usually guarantee optimal regret rates in both adversarial and stochastic settings, without being aware of which environment they are in. 
Various work study the case of bandits without budget constraints (see, \emph{e.g.}, \citet{bubeck2012best,seldin2014one,Auer-colt16}).
\citet{rangi2018unifying} present an algorithm for the BwK and adversarial BwK setting for the special case when there is only one constrained resource, including time (as mentioned, this particular setting is simpler and admits much stronger performance guarantees).
In the context of online allocation problems with fixed per-iteration budget,~\citet{balseiro2020dual,balseiro2022best} propose a class of algorithms which attain asymptotically optimal performance in the stochastic case, and they attain an asymptotically optimal constant-factor competitive ratio when the input is adversarial. In their setting, as we already mentioned, at each round the input $(f_t,c_t)$ is observed by the decision maker \emph{before} they make a decision. This makes the problem essentially different from ours.
In the specific  \emph{AdWords} setting, \citet{mirrokni2012simultaneous} present an algorithm which guarantees optimal competitive ratio under adversarial inputs, and an improved competitive ratio for stochastic inputs.

\xhdr{Other related work.} 
\citet{sankararaman2021bandits} show that achieving $O(\log T)$ regret for BwK is possible if and only if there is only one resource, and the best distribution over arms
reduces to the best fixed arm. If either of these conditions fail, one cannot hope for a better rate than $\Omega(T^{1/2})$.
In the simplified setting with only one constrained resource and unlimited number of rounds. it is possible to obtain instance-dependent $\textnormal{polylog}(T)$ regret bounds under various assumptions (see, e.g.,  \citet{Gyorgy-ijcai07,tran2010epsilon,tran2012knapsack}).
\citet{wu2015algorithms} assume to have a single resource, and that resource consumption is deterministic.  
Another line of related work concerns online convex optimization with constraints (see, \emph{e.g.}, \citet{mahdavi2012trading,mahdavi2013stochastic,chen2017online,neely2017online,chen2018bandit}), where it is usually assumed that the action set is a convex subset of $\R^m$, in each round rewards (resp., costs) are concave (resp., convex), and, most importantly, resource constraints only apply at the last round, while in BwK the budget constraints hold for all rounds.

\xhdr{Concurrent work.} The work by \citet{fikioris2023approximately} introduces the notion of \emph{Approximately Stationary Bandits with Knapsacks}, which provides a way to interpolate between the fully stochastic and the fully adversarial setting. Based on the conference version of this paper, the authors extend the results for our algorithm to the case where the fluctuations in expected rewards and
consumptions of any arm are limited across rounds. Based on this assumption they provide algorithms guaranteeing a competitive ratio which smoothly transitions between the best guarantees in the two extreme cases.
We view the results of \cref{sec:nonstationary} as complementary to those of \citet{fikioris2023approximately}.

%% file: preliminaries.tex
\section{Preliminaries}\label{sec: preliminaries}

We denote vectors by bold fonts. Given vector $\vx$, let $\vx[i]$ be its $i$-th component. The set $\{1,\ldots,n\}$, with $n\in\N_{>0}$, is compactly denoted as $[n]$. %
Given a discrete set $S$, we denote by $\Delta^{S}$ the $|S|$-simplex.
Finally, given a proposition $\term{p}$, we denote with $\indicator{\term{p}}$ the indicator function of that proposition: $\indicator{\term{p}}=1$ if $\term{p}$ is true, and $\indicator{\term{p}}=0$ otherwise.

\xhdr{Basic Setup.} 
There are $T$ rounds and $m$ resources. A decision maker has a non-empty set of available strategies $\cX\subseteq\R^n$ (this set may be non-convex, integral, and even non-compact).
In each round $t\in[T]$, the decision maker chooses $\vx_t\in\cX$, and subsequently observes a reward function $f_t:\cX\to[0,1]$, and a function $c_t:\cX\to[0,1]^{m}$ specifying resources consumption (both $f_t$ and $c_t$ need not be convex). 
Each resource $i\in[m]$ is endowed with a budget $B_i$ to be spent over the $T$ steps. Since $c_t(\vx)[i]\ge0$, for all $t,i$, and $\vx$, budgets cannot be replenished. We denote by $\vrho\defeq(\rho_1,\ldots,\rho_m)\in\R_{>0}^m$ the vector of per-iteration budgets, where for each $i\in[m]$ we have $B_i=T\rho_i$. Without loss of generality we let $\rho_1=\ldots=\rho_m=\rho$, and $B_1=B_2=\ldots=B_m=B$.
A problem with arbitrary budgets can be reduced to this setting by dividing, for each resource $i\in[m]$, all per-round resource consumption $c_t(\cdot)[i]$ by $B_i / \min_j B_j$. 
We focus on the regime $B=\Omega(T)$, and we study two feedback models: \emph{bandit feedback} (no auxiliary feedback other than $f_t(\vx_t),c_t(\vx_t)$ is observed by the decision maker), and \emph{full feedback} ($f_t,c_t$ are observed).
Let $\gamma_t\defeq(f_t,c_t)$, and $\vgamma_{T}\defeq(\gamma_t)_{t=1}^T$ be the sequence of inputs up to time $T$. 
At each step $t$, the decision maker can condition their decision on $\vgamma_{t-1}$, and on the sequence of prior decisions $\vx_1,\ldots,\vx_{t-1}$, but no information about future rewards or resource consumption is available. The repeated decision making process stops at any round $\tau\le T$ in which the total consumption of any resource $i$ exceeds its budget $B_i$. The goal of the decision maker is to maximize its total reward. 
Following previous work \citep{badanidiyuru2013bandits,agrawal2014bandits,immorlica2019adversarial},
we assume there exists a \emph{void action} $\nullx\in\cX$ with reward 0, and such that $c_t(\nullx)[i]=0$ for all resources $i$. This guarantees the existence of a feasible solution (i.e., a sequence of decisions which do not violate resource constraints).

\xhdr{Regret Minimization.} 
A regret minimizer for an arbitrary set $\cW$ is an abstract model for a decision maker that repeatedly interacts with a black-box environment. 
At each time $t$, the regret minimizer can perform two operations: (i) \nextelement: this procedure outputs an element $\vw_t \in \cW$; (ii) \observeutility: this procedure updates the internal state of the regret minimizer using the environment's feedback, in the form of a utility function $\ell_t:\cW\to\R$. The utility function can depend adversarially on the  sequence of outputs $\vw_1,\ldots,\vw_{t-1}.$
The decision making process encoded by the regret minimizer is \emph{online}: at each time $t$, the output of the regret minimizer can depend on the sequence $(\vw_{t'},\ell_{t'})_{t'=1}^{t-1}$, but no information about future utilities is available. 
The objective of the regret minimizer is to output a sequence of points in $\cW$ so that the \emph{cumulative regret}
$R^T\defeq \sup_{\vw^\ast\in\cW}\sum_{t=1}^T\mleft(\ell_t(\vw^\ast)-\ell_{t}(\vw_t)\mright)$
grows asymptotically sublinearly in the time $T$.
For a review of the various regret minimizers available for the full and bandit feedback setting see~\citet{cesa2006prediction}.

%% file: mixtures.tex
\section{Strategy Mixtures}\label{sec: mixtures}

We will need to work with the set of probability measures on the Borel sets of $\cX$. We refer to this set as the set of \emph{strategy mixtures} and denote it as $\Xi$.
We endow $\cX$ with the Lebesgue $\sigma$-algebra. 
We assume that all possible functions $f_t,c_t$ are measurable with respect to every probability measure $\vxi \in \Xi$. This ensures that the various expectations taken are well-defined, since the functions are assumed to be bounded above, and are therefore integrable.

It is well-known that the Dirac measures $\delta_x$ for $x\in \cX$ which assign 1 to a set $\cA \subseteq \cX$ if and only if $x\in \cA$ form the extreme points of the convex set of strategy mixtures $\Xi$.
The Dirac mass $\delta_\nullx$ is the strategy that deterministically plays the void action. We define $\vxi_{\nullx}\defeq \delta_\nullx$.

\subsection{On Strong Duality}

Given two arbitrary measurable functions $f:\cX\to[0,1]$, $c:\cX\to[0,1]^m$, we define the following linear program, which chooses the strategy $\vxi$  that maximizes the reward $f$, while keeping the expected consumption of every resource $i\in[m]$ given $c$ below the target $\rho$:
\begin{equation}\label{eq:opt lp gen}
    \OPTLP_{f,c}\defeq\mleft\{\hspace{-1.25mm}\begin{array}{l}
        \displaystyle
        \sup_{\vxi\in\Xi}\E_{\vx\sim\vxi}\mleft[f(\vx)\mright] \\ [2mm]
        \displaystyle \text{\normalfont s.t. } \E_{\vx\sim\vxi}\mleft[ c(\vx)\mright]\le \vrho
    \end{array}\mright.,
\end{equation}
where $\E_{\vx\sim\vxi}\mleft[c_t(\vx)\mright]=(\E_{\vx\sim\vxi}\mleft[c_t(\vx)[i]\mright])_{i=1}^m\in[0,1]^m$.

By letting $\cI$ be an arbitrary set of possible input pairs $(f,c)$, the Lagrangian relaxation of LP \eqref{eq:opt lp gen} is defined as follows.
\begin{definition}[Lagrangian Function]\label{def:lagrangian}
The \emph{Lagrangian function} $L:\Xi\times\R_{\ge 0}^m \times \cI \to\R$ is such that, for any $\vxi\in\Xi$, $\vlambda\in\R_{\ge 0}^m$, $(f,c)\in\cI$ it holds
\[
L(\vxi,\vlambda, f, c)\defeq \E_{\vx\sim\vxi}\mleft[ f(\vx)\mright] + \langle\vlambda,\vrho-\E_{\vx\sim\vxi}\mleft[ c(\vx)\mright]\rangle.
\]
\end{definition}

Next, we show that, when the decision maker is allowed to choose a strategy mixture, we can recover strong duality even if $f$ and $c$ are arbitrary non-convex functions (omitted proofs can be found in \cref{sec:omitted proofs}).
\begin{theorem}\label{strong duality}
	Let $f:\cX\to[0,1]$, $c:\cX\to[0,1]^m$, $(f,c)\in\cI$. It holds:
	\[
	\sup_{\vxi\in\Xi }\inf_{\vlambda\ge 0} L(\vxi,\vlambda,f, c)=\inf_{\vlambda\ge 0}\sup_{\vxi\in\Xi } L(\vxi,\vlambda, f, c)=\OPTLP_{f,c}.
	\]
\end{theorem}
This theorem can be derived from \citet[Theorem 1, \S8.6]{luenberger1997optimization}. For completeness and ease of readability, we give a proof specific to our setting.
The proof is based on standard convex-optimization arguments. In general, a semi-infinite linear optimization problem does not admit strong duality (see the example in \cref{appendix strong duality}).
However, the existence of a strategy mixture $\vxi_\nullx$ corresponding to playing deterministically the void action $\nullx$ allows us to follow closely the standard proof of strong duality via Slater's condition, since it yields the existence of a strictly feasible solution.

Then, we show that we can restrict the set of admissible dual vectors to
\begin{equation}\label{eq:dual set}
    \cD\defeq\mleft\{\vlambda\in\R_{\ge0}:\|\vlambda\|_1\le1/\rho\mright\},
\end{equation} while continuing to satisfy strong duality.

\begin{lemma}\label{lemma:strong duality 2}
	Let $\cD$ be defined as in \cref{eq:dual set}. Given $f:\cX\to[0,1]$, $c:\cX\to[0,1]^m$, $(f,c)\in\cI$, it holds
	\[
	\hspace{-.2cm}
	\sup_{\vxi\in\Xi }\inf_{\vlambda\in\cD} L(\vxi,\vlambda, f, c)=\inf_{\vlambda\in\cD}\sup_{\vxi\in\Xi } L(\vxi,\vlambda, f, c)=\OPTLP_{ f, c}.
	\]
\end{lemma}

From \cref{lemma:strong duality 2} we have that $\vlambda$ is chosen from a compact set.
By noting that the supremum over a set of lower semicontinuous (LSC) functions is LSC~\citep[Lemma 2.41]{aliprantis2006infinite}, we get that $\sup_{\vxi\in\Xi } L(\vxi,\vlambda,f,c)$ is LSC as a function of $\vlambda$.
It follows by a generalization of Weierstrass' theorem that an optimal $\vlambda^*$ exists~\citep[Theorem 2.43]{aliprantis2006infinite}.
Therefore, going forward, we will replace all infima over $\cD$ by minima when needed.

\subsection{Baselines}

In this section we provide details on the baselines for the case of adversarial and stochastic inputs, respectively.

\xhdr{Baseline adversarial setting.}
Given a sequence of inputs $\vgamma_T$, the baseline for the adversarial setting is the total expected reward of the best \emph{fixed} policy in $\Xi$, such that strategies are drawn from the same fixed mixture until the budget is fully depleted, and the void action is selected afterwards. Following the notation of \citet{immorlica2019adversarial}, we denote its value by $\OPTFD_{\vgamma}$. Given $\tau\in[T]$, we write $\OPTFD_{\vgamma,\tau}$ to denote the expected reward of the best fixed policy for inputs restricted to $(\gamma_1,\ldots,\gamma_\tau)$.
Moreover, for any sequence of inputs $\vgamma_T$, and for any $\tau\in[T]$, let $\tilde f_\tau: \cX\to[0,1]$ and $\tilde c_\tau:\cX\to [0,1]^m$ be such that, for each $\vx\in\cX$:
\begin{equation}\label{eq: tilde functions}
\tilde f_\tau(\vx)\defeq \frac{1}{\tau}\sum_{t=1}^\tau f_t(\vx)\hspace{.2cm}\textnormal{and}\hspace{.2cm}\tilde c_\tau(\vx)\defeq \frac{1}{\tau}\sum_{t=1}^\tau c_t(\vx).
\end{equation}
Then, for $\tau\in[T]$, we define $\OPTLP_{\tilde f_\tau,\tilde c_\tau}$ according to \cref{eq:opt lp gen}. The value of these LPs will be essential during the regret analysis.

\xhdr{Baseline stochastic setting.} 
In the stochastic version of the problem, each input $\gamma_t=(f_t,c_t)$ is drawn i.i.d. from some unknown distribution $\distr$ over a set of possible input pairs $\inputset$.
Let $\bar f:\cX\to[0,1]$ be the expected reward function, and $\bar c:\cX\to[0,1]^m$ be the expected resource-consumption function (where both expectations are taken w.r.t. $\distr$). 
Let $\Psi$ be the set of dynamic policies specifying the current strategy mixture $\vxi_t$ as a function of the past history.
The baseline for the stochastic setting $\OPTDP$ is given by $\OPTDP\defeq \sup_{\psi \in \Psi} \E_{\vgamma\sim\distr}[\OPTDP_{\psi,\vgamma}]$, where $\OPTDP_{\psi,\vgamma}$ is the value of the policy $\psi$ under the sequence of inputs $\vgamma$. Intuitively, $\OPTDP$ is the value of the best dynamic policy when the decision maker knows $\distr$, but gets to observe the realized $\gamma_t$ only after having made the decision at $t$.
In the following, we use the solution to LP~\eqref{eq:opt lp gen} initialized with reward function $\bar f$, and cost function $\bar c$, as an upper bound to the value of the optimal policy $\OPTDP$.
In particular, we prove the following.
\begin{lemma}\label{lemma: stoc opt ub}
	Given a distribution over inputs $\distr$, let $\bar f:\cX\to[0,1]$ be the expected reward function, and $\bar c:\cX\to[0,1]^m$ be the expected resource-consumption function. Then, $T \cdot \OPTLPE\ge\OPTDP$.
\end{lemma}

%% file: alg.tex
\section{Meta-Algorithm}

Our framework assumes access to two regret minimizers with the following characteristics. The first one, which we denote by $\cRp$, is the \emph{primal regret minimizer} which outputs strategy mixtures in $\Xi$, and receives as feedback the linear utility $\lossp:\Xi\to\R$ such that, for each $\vxi\in\Xi$, $\lossp(\vxi)\defeq \E_{\vx\sim\vxi}\mleft[f_t(\vx)\mright] - \langle\vlambda_t,\E_{\vx\sim\vxi}\mleft[c_t(\vx)\mright]\rangle$.
The second regret minimizer, which we denote by $\cRd$, is the \emph{dual regret minimizer}, and it outputs points in the space of dual variables $\cD$. Moreover, $\cRd$ receives as feedback the linear utility $\lossd:\cD\to\R$ such that, for each $\vlambda\in\cD$, $\lossd(\vlambda)\defeq -\langle\vlambda, \vrho-\E_{\vx\sim\vxi_t}[c_t(\vx)]\rangle$.
The primal regret minimizer $\cRp$ may have either bandit or full feedback, depending on the setting of interest. The dual regret minimizer $\cRd$ has full feedback by construction.
Finally, we denote by $\cump$ (resp., $\cumd$) the upper bound on the cumulative regret guaranteed by $\cRp$ (resp., $\cRd$).

\begin{remark}\label{remark:omd}
    In order to guarantee convergence in both the stochastic and adversarial setting, it will be enough to set $\cD\defeq\{\vlambda\in\R_{\ge0}:\|\vlambda\|_1\le1/\rho\}$ (see \cref{sec:adv,sec:stoc}). Therefore, a natural choice for the dual regret minimizing algorithm $\cRd$ is, for example, \emph{online mirror descent} (OMD) with negative entropy as reference function, which guarantees a regret upper bound of $\cumd=O(1/\rho\sqrt{T\log(m+1)})$ \citep{nemirovskij1983problem,beck2003mirror}.
\end{remark}

\cref{alg:meta alg} summarizes the structure of our meta-algorithm. For each $t$, the meta-algorithm first computes a primal and dual decision through $\cRp$ and $\cRd$, respectively (see the invocation of $\nextelement$). 
The action played by the decision maker at $t$ is going to be $\vx_t\sim\vxi_t$. Then, $(f_t,c_t)$ are observed, and the budget consumption is updated according to the realized cost vector $c_t(\vx_t)$. 
Finally, the internal state of the two regret minimizer is updated according to the feedback specified by $\lossp,\lossd$ (see the invocation of $\observeutility[]$). Notice that the primal regret minimizer $\cRp$ may as well receive partial feedback (\emph{i.e.}, observe only $(f_t(\vx_t),c_t(\vx_t))$ for each $t$). In the following sections we show that, with an appropriate choice of $\cRp$, \cref{alg:meta alg} can be adapted also to this setting.
The algorithm terminates when the agent has no sufficient budget, or when the time horizon $T$ is reached.

\begin{algorithm}[tb]
	\caption{Meta-algorithm for strategy mixture $\Xi$.}
	\label{alg:meta alg}
	\KwData{parameters $B,T$, primal regret minimizer $\cRp$, dual regret minimizer $\cRd$}
	{\bfseries Initialization:} $\forall i\in[m], B_{i,1}\gets B, \,\vrho\gets \vone\cdot B/T$, and initialize $\cRp,\cRd$\\
	
	\For{$t = 1, 2, \ldots , T$}{
		{\bfseries Primal decision:} $\Xi\ni\vxi_t\gets \cRp.\nextelement$,
		\[
		\vx_t\gets \mleft\{\hspace{-1.25mm}\begin{array}{l}
			\displaystyle
			\vx\sim\vxi_t \hspace{.5cm}\text{\normalfont if } B_{i,t} \ge 1,\, \forall i\in[m]\\ [2mm]
			\nullx \hspace{.5cm}\text{otherwise}
		\end{array}\mright.
		\]
		{\bfseries Dual decision:} $\cD\ni \vlambda_t\gets\cRd.\nextelement$\\
		
		{\bfseries Observe request:} 
		\begin{OneLiners}
			\item Observe $(f_t,c_t)$
			\item Update available resources: $B_{i,t+1}\gets B_{i,t} - c_t(\vx_t)[i],\,\forall i\in[m]$
		\end{OneLiners}
		
		{\bfseries Primal update:} 
		\begin{OneLiners}
			\item  $\lossp\gets$ linear utility defined as $\quad\lossp:\Xi\ni\vxi\mapsto \E_{\vx\sim\vxi}\mleft[f_t(\vx)\mright] - \langle\vlambda_t,\E_{\vx\sim\vxi}\mleft[c_t(\vx)\mright]\rangle$
			\item  $\cRp.\observeutility[\lossp]$
		\end{OneLiners}
	
		{\bfseries Dual update:} 
		\begin{OneLiners}
			\item $\lossd\gets$ linear utility defined as  $\quad\lossd:\cD\ni\vlambda\mapsto -\langle\vlambda, \vrho-\E_{\vx\sim\vxi_t}[c_t(\vx)]\rangle$
			\item $\cRd.\observeutility[\lossd]$
		\end{OneLiners}
		 
	}
\end{algorithm}

\begin{remark}\label{remark:sequential}
    In Algorithm~\ref{alg:meta alg} we let the primal and dual regret minimizer choose their decisions simultaneously. However, another possibility is to let the regret minimizers choose the decisions sequentially. In particular, we could let the dual regret minimizer select a dual variable $\vlambda_t$, and then let the primal regret minimizer observe $\vlambda_t$ and choose a decision. The only properties that we need in order to prove our results  are the bounds on the primal and dual regret. Hence, when the regret minimizers play sequentially, we need a dual regret minimizer that has no-regret guarantees even if the primal observes the dual decision.
    A dual regret minimizer that has no-regret with respect to an almighty adversary (i.e., an adversary who has access to the realization of any random bits employed by the algorithm) satisfies this property. 
    All the dual regret minimizers that we consider in this work (e.g., mirror descent) do not employ randomization, and hence they have no regret against an almighty adversary.
\end{remark}

%% file: adversarial.tex
\section{Regret Bound for the Adversarial Setting}
\label{sec:adv}

In this section, we assume that, for each $t\in[T]$, the request $(f_t,c_t)$ is chosen by an oblivious adversary, and we look at the worst-case performance over all possible inputs. We show that, in this setting, Algorithm~\ref{alg:meta alg} is $\alpha$-competitive, with $\alpha\defeq 1/\rho$. This is, to the best of our knowledge, the first constant-factor competitive ratio for the adversarial setting in the $B=\Omega(T)$ regime.
The competitive ratio $\alpha$, being defined as the maximum cost (\emph{i.e.}, $1$ in our setting) over the per-round budget, captures the \emph{relative wealthiness} of the decision maker. As one may expect, bidding strategies may perform poorly when budgets are small compared to the costs, but better performance can be guaranteed with bigger budgets.
In particular, we provide the following convergence guarantees of Meta-Algorithm~\ref{alg:meta alg} in the adversarial setting.

\begin{theorem}\label{thm:adversarial}
Consider Meta-Algorithm~\ref{alg:meta alg} equipped with two arbitrary regret minimizers $\cRp$ and $\cRd$ for the sets $\Xi$ and $\cD$, respectively. Suppose they guarantee cumulative regret up to time $T$ which is upper bounded by $\cump$ and $\cumd$, respectively. Suppose requests are chosen by an oblivious adversary. Letting $\alpha\defeq 1/\rho$, for each $\delta>0$ we have
\[
\OPTFD_{\vgamma} -\alpha\,\REW_{\vgamma}\le O(\alpha^2\sqrt{T\ln(T/\delta)})+\cump[T]+ \cumd[T]
\]
with probability at least $1-\delta$, where $\REW_{\vgamma}\defeq\sum_{t}  f_t(\vx_t)$ is the reward of the algorithm for the sequence of inputs $\vgamma$.
\end{theorem}

Intuitively, the  proof works as follows: the algorithm depletes the budget close to the end of the time horizon $T$. This is because the primal regret minimizer incurs a cost of $T-\tau$ for stopping too early, where $\tau$ is the stopping time of the algorithm (see \cref{eq:stopping time}). 
Since we know that $\tau$ must be close to $T$, we use as an intermediate baseline the optimal fixed action of the unconstrained problem over time horizon $\tau$ (i.e., the baseline just maximizes $\sum_{t=1}^\tau f_t(\vx)$, without any resource constraint). This allows us to avoid guessing the exact stopping time as it happens in the algorithm by \citet{immorlica2019adversarial}, since the  unconstrained optimum never fully depletes the resources. 
Finally, we prove that our algorithm over $\tau$ rounds provides a $B/T$-approximation of the unconstrained baseline (see \cref{eq: bound a}). 

\begin{proof}{Proof of \cref{thm:adversarial}.} Let $\tau$ be the stopping time of Algorithm~\ref{alg:meta alg}, i.e. when $B_{i,t} < 1$. We proceed in three steps.

\textbf{Step 1: lower bound on the reward up to $\tau$.} 
First, we provide a lower bound on the reward guaranteed by Algorithm~\ref{alg:meta alg} up to the stopping time $\tau$.
The cumulative external regret of $\cRp$ up to the stopping time $\tau$ is
\[
\regp[\tau]=\sup_{\vxi\in\Xi}\sum_{t=1}^\tau\mleft(\lossp(\vxi) - \lossp(\vxi_t)\mright)\le \cump[\tau].
\]
By definition of $\lossp$, we have
\begin{align}
& \sup_{\vxi\in\Xi}\sum_{t=1}^\tau\mleft(\E_{\vx\sim\vxi}\mleft[f_t(\vx)\mright] - \langle\vlambda_t,\E_{\vx\sim\vxi}\mleft[c_t(\vx)\mright]\rangle - \E_{\vx\sim\vxi_t}\mleft[f_t(\vx)\mright]+\langle\vlambda_t,\E_{\vx\sim\vxi_t}\mleft[c_t(\vx)\mright]\rangle\mright)\le \cump[\tau]. \nonumber
\end{align}
Then, by rearranging, 
\begin{equation}\label{eq:adv 1}
    \sum_{t=1}^\tau \E_{\vx\sim\vxi_t}\mleft[f_t(\vx)\mright]\ge \sup_{\vxi\in\Xi}\sum_{t=1}^\tau\mleft(\E_{\vx\sim\vxi}\mleft[f_t(\vx)\mright] - \langle\vlambda_t,\E_{\vx\sim\vxi}[c_t(\vx)]\rangle + \langle\vlambda_t,\E_{\vx\sim\vxi_t}\mleft[c_t(\vx_t)\mright]\rangle\mright)-\cump[\tau].
\end{equation}
By definition of dual regret minimizer $\cRd$, for any $\vlambda\in\cD$ (we will specify a precise value for $\vlambda$ in the final step of the proof), we have $\sum_{t=1}^\tau\mleft(\lossd (\vlambda)-\lossd (\vlambda_t)\mright)\le\cumd[\tau]$. Then, by definition of $\lossd$,
\begin{equation*}
    \sum_{t=1}^\tau\langle\vlambda_t,\E_{\vx\sim\vxi_t}[c_t(\vx)]\rangle\ge \sum_{t=1}^\tau \mleft(\langle\vlambda_t,\vrho\rangle - \langle\vlambda,\vrho\rangle+\langle\vlambda,\E_{\vx\sim\vxi_t}[c_t(\vx)]\rangle\mright)- \cumd[\tau].
\end{equation*}
By substituting in Equation~\eqref{eq:adv 1},
\begin{align}
    \sum_{t=1}^\tau \E_{\vx\sim\vxi_t}\mleft[f_t(\vx)\mright]& \ge -\cump[\tau]- \cumd[\tau] +\sup_{\vxi\in\Xi}\sum_{t=1}^\tau\mleft(\E_{\vx\sim\vxi}\mleft[f_t(\vx)\mright]+ \langle\vlambda_t,\vrho-\E_{\vx\sim\vxi}[c_t(\vx)]\rangle  
    - \langle\vlambda,\vrho-\E_{\vx\sim\vxi_t}[c_t(\vx)]\rangle\mright).\label{eq: bound sum f}
\end{align}
Then, we bound the term  
\[\circled{A}\defeq \sup_{\vxi\in\Xi}\sum_{t=1}^\tau\mleft(\E_{\vx\sim\vxi}\mleft[f_t(\vx)\mright]+ \langle\vlambda_t,\vrho-\E_{\vx\sim\vxi}[c_t(\vx)]\rangle\mright).\]
Let $\OPTSTAR\defeq\sup_{\vx \in \cX} \sum_{t=1}^\tau f_t(\vx)$, that is, $\OPTSTAR$ is the supremum of the unconstrained problem.
Then, for each $\epsilon>0$, there exists an $\vx^\ast \in \cX$ such that $\sum_t f_t(\vx^\ast)\ge \OPTSTAR-\epsilon$.
Then, we show that 
\begin{align}
\circled{A} & \ge \max_{\vx\in\{\vx^\ast,\nullx\}} \sum_{t=1}^\tau\mleft(\E_{\vx\sim\vxi}\mleft[f_t(\vx)\mright]+ \langle\vlambda_t,\vrho-\E_{\vx\sim\vxi}[c_t(\vx)]\rangle\mright)\nonumber
\\&\ge \rho \,\OPTSTAR-\epsilon.\label{eq: bound a}
\end{align}
To do so, we consider two cases. First, if $\sum_{t=1}^\tau f_t(\vx^\ast)\ge \sum_{t=1}^\tau \langle \vlambda_t, c_t(\vx^\ast)\rangle$, then the value of the function for $\vx^\ast$ is at least  
\begin{align*}
\circled{A} & \ge \sum_{t=1}^\tau\mleft(f_t(\vx^\ast)+ \langle\vlambda_t,\vrho-c_t(\vx^\ast)\rangle\mright) 
\\&\ge
\sum_{t=1}^\tau\mleft(f_t(\vx^\ast)+ \langle\vlambda_t,\rho \cdot c_t(\vx^\ast)-c_t(\vx^\ast)\rangle\mright)
\\ & \ge 
\sum_{t=1}^\tau f_t(\vx^\ast)-(1-\rho) \sum_{t=1}^\tau \langle\vlambda_t,c_t(\vx^\ast)\rangle 
\\ & \ge
\rho \sum_{t =1}^\tau f_t(\vx^\ast) = \rho\,\OPTSTAR -\epsilon,
\end{align*}
where the second inequality holds since $c_t(\cdot)\in [0,1]^m$, for each $t\in[T]$.
Otherwise, if $\sum_{t} f_t(\vx^\ast)< \sum_{t} \langle \vlambda_t, c_t(\vx^\ast)\rangle$, we have that the null action $\nullx$ has value at least  
\begin{align*}
\circled{A}\ge\sum_{t=1}^\tau \langle\vlambda_t,\vrho\rangle \ge \sum_{t=1}^\tau \langle\vlambda_t,\rho\cdot c_t(\vx^\ast)\rangle \ge
\rho\,\sum_{t =1}^\tau f_t(\vx^\ast)\ge \OPTSTAR-\epsilon.
\end{align*}
This shows that \cref{eq: bound a} holds. Hence, $\circled{A}\ge \rho\,\OPTSTAR$.
Then, by substituting this in \cref{eq: bound sum f}, we have
\begin{align}
    \sum_{t=1}^\tau \mleft(\E_{\vx\sim\vxi_t}\mleft[f_t(\vx)\mright]-\langle\vlambda,\E_{\vx\sim\vxi_t}[c_t(\vx)]\rangle\mright)
    \ge -\cump[\tau]- \cumd[\tau]+\rho\OPTSTAR - \tau\, \langle\vlambda,\vrho\rangle.\label{eq: adv Ef Ec}
\end{align}

\textbf{Step 2: relating expectations and their realizations.} Now, we have to relate the lefthand side of the above inequality to its realized value $\sum_{t=1}^\tau\mleft(f_t(\vx_t)-\langle\vlambda,c_t(\vx_t)\mright)$. In order to do this, let
\[
W_t\defeq -f_t(\vx_t)+\E_{\vx\sim\vxi_t}[f_t(\vx)]+\langle \vlambda,c_t(\vx_t)-\E_{\vx\sim\vxi_t}[c_t(\vx)]\rangle,
\]
and observe that $W_1,\ldots,W_\tau$ is a martingale difference sequence, with $|W_t|\le 1+\|\vlambda\|_\infty$. Then, by the Azuma-Hoeffding inequality, we have that, for any $\tau\in[T]$,
\[
\Pr\mleft[\sum_{t=1}^\tau W_t > \mleft(1+\|\vlambda\|_{\infty}\mright)\sqrt{2\tau\ln(1/\delta)}\mright]\le\delta.
\]
Let, $q(\tau,\vlambda,\delta)\defeq \mleft(1+\|\vlambda\|_{\infty}\mright)\sqrt{2\tau\ln(1/\delta)}$.
Then, by taking a union bound, we have
\[
\Pr\mleft[\forall\tau\in[T],\,\,\sum_{t=1}^\tau W_t \le q(\tau,\vlambda,\delta)\mright]\ge 1-T\,\delta.
\]
Therefore, for any $\delta>0$, with probability at least $1-\delta$, 
\begin{align}
\sum_{t=1}^\tau\mleft(f_t(\vx_t)-\langle\vlambda,c_t(\vx_t)\rangle\mright)&\ge -q(\tau,\vlambda,\delta/T) 
+\sum_{t=1}^\tau\mleft(\E_{\vx\sim\vxi_t}[f_t(\vx)]-\langle\vlambda,\E_{\vx\sim\vxi_t}[c_t(\vx)]\rangle\mright)\nonumber
\nonumber \\ &
\ge -\underbrace{\mleft(q(\tau,\vlambda,\delta/T) +\cump[\tau]+ \cumd[\tau]\mright)}_{\circled{B}}+\rho\,\OPTSTAR - \tau\, \langle\vlambda,\vrho\rangle,\nonumber
\end{align}
where the last inequality is by \cref{eq: adv Ef Ec}. 

\textbf{Step 3: putting everything together.} First, we rewrite $\OPTSTAR$ as a function of the baseline $\OPTFD_{\vgamma}$. In particular, we have
\[
\rho\,\OPTSTAR \ge \rho\, \OPTFD_{\vgamma,\tau} \ge\rho\,\mleft(\OPTFD_{\vgamma}-T+\tau\mright).
\]
By definition $\REW_{\vgamma}\defeq\sum_{t=1}^\tau(f_t(\vx_t))$. Then, 
\begin{align}
\REW_{\vgamma} & \ge \rho\,\OPTSTAR-\sum_{t=1}^\tau\langle\vlambda,\vrho-c_t(\vx_t)\rangle -\circled{B}\ge \frac{\OPTFD_{\vgamma}-T+\tau}{\alpha}-\sum_{t=1}^\tau\langle\vlambda,\vrho-c_t(\vx_t)\rangle -\circled{B}.\label{eq:lb rew}
\end{align}

If $\tau=T$ (\emph{i.e.}, the stopping time coincides with the time horizon $T$), in order to get the result it is enough to set $\vlambda=\vec{0}$, and to substitute the above expression in the definition of regret. 
Otherwise, if $\tau<T$, it means that there exists a resource $\stopi\in [m]$ for which 
\begin{equation}\label{eq:stopping time}
\sum_{t=1}^\tau c_t(\vx_t)[\stopi]+1 \ge \rho T,
\end{equation}
where, in our setting, $1$ is the maximum observable cost.
Then, we set $\vlambda$ as follows: $\vlambda[\stopi]=1/\rho$, and $\vlambda[i]=0$ for all $i\ne\stopi$. For this choice of $\vlambda$, and by exploiting \cref{eq:stopping time}, we have
\begin{align*}
\sum_{t=1}^\tau\langle\vlambda,\vrho-c_t(\vx_t)\rangle & =\alpha\sum_{t=1}^\tau\mleft(\rho-c_t(\vx_t)[\stopi]\mright)\\
& \le \tau - T +\alpha.
\end{align*}
Then, by substituting the above expression in \cref{eq:lb rew},
\[
\REW_{\vgamma}\ge \frac{\OPTFD_{\vgamma}-T+\tau}{\alpha}- (\tau-T)-\alpha -\circled{B}.
\]
Finally, we have
\begin{align*}
\OPTFD_{\vgamma} -\alpha\,\REW_{\vgamma} & \le (T-\tau)- \alpha(T-\tau)+\alpha^2 +\alpha\cdot\circled{B}\\
& \hspace{-.7cm}\le \alpha^2+\alpha\mleft(q(\tau,\vlambda,\delta/T) +\cump[\tau]+ \cumd[\tau]\mright)\\
& \hspace{-.7cm}\le \alpha^2 +\alpha(1+\alpha)\sqrt{2T\ln(T/\delta)}+\cump[T]+ \cumd[T],
\end{align*}
where the last inequality holds because the error terms are increasing in $t$. This concludes the proof.
\end{proof}

\begin{remark}\label{remark:regret min x xi}
Let $\hat\lossp:\cX\ni\vx\mapsto f_t(\vx)-\langle\vlambda_t,c_t(\vx)\rangle$. Then, by definition of the set of strategy mixtures $\Xi$ (see \cref{sec: mixtures}), for each $\tau \in [T]$ it holds  \[\sup_{\vx\in\cX}\sum_{t=1}^\tau\hat\lossp(\vx)=\sup_{\vxi\in\Xi}\sum_{t=1}^\tau\lossp(\vxi).\]
\end{remark}

\begin{remark}\label{remark:regret high prob}
The guarantees of \cref{thm:adversarial} can be extended, with minor modifications, to the bandit feedback setting. In particular, let $\hat\lossp:\cX\to\R$ be defined as in \cref{remark:regret min x xi}, and $\cRp$ be a primal regret minimizer guaranteeing, with probability at least $1-\delta$, that
$
\sup_{\vx\in\cX}\sum_{t=1}^\tau\mleft(\hat\lossp(\vx)-\hat\lossp(\vx_t)\mright)\le \cump[\tau,\delta].
$
By \cref{remark:regret min x xi}, the above high-probability regret bound implies that, with probability at least $1-2\delta$, $\regp[\tau]\le\cump[\tau,\delta]+ O(\sqrt{T \ln(T/\delta)}/\rho)$ (see \cref{sec:appAdv}). Then, it is possible to follow the proof of \cref{thm:adversarial}, and to recover, via the application of an additional union bound, the same guarantees with probability at least $1-3\delta$.
\end{remark}

%% file: stochastic.tex
\section{Regret Bound for the Stochastic Setting}
\label{sec:stoc}

In this section, we prove an optimal regret upper bound matching that of \citet{badanidiyuru2013bandits,immorlica2019adversarial}. We employ the notion of \emph{expected Lagrangian game} employed by \citet{immorlica2019adversarial}. However, by working in the space of strategy mixtures $\Xi$, we provide a simplified analysis.
In particular, working on the mixtures $\Xi$, we can provide deterministic bounds on the regret, while \citet{immorlica2019adversarial} works with regret minimizers in high probability.
Moreover, our approach allows to generalize the result of \citet{immorlica2019adversarial} to general, \emph{e.g.}, non-convex, problems.

First, we introduce the following preliminary lemma.
\begin{lemma}\label{lemma:expected game}
Given $\tau\in[T]$, for $\delta\in(0,1)$, with probability at least $1-2\delta$, the average strategy mixture $\bar\vxi\in\Xi$ up to $\tau$ is such that, for any $\vlambda\in\cD$,   
$
\bar L(\bar\vxi,\vlambda)\ge  \OPTLPE- \frac{1}{\tau}\mleft(\cump+\cumd+4(1+1/\rho)\sqrt{2T\log(T/\delta)}\mright).
$
\end{lemma}

Now, we prove the main theorem.
\begin{theorem}\label{thm:stochastic}
	Consider Meta-Algorithm~\ref{alg:meta alg} equipped with two arbitrary regret minimizers $\cRp$ and $\cRd$ for the sets $\Xi$ and $\cD$, respectively. In particular, assume that they guarantee a cumulative regret up to time $T$ which is upper bounded by $\cump$ and $\cumd$, respectively. For each $t\in[T]$, let the inputs$(f_t,c_t)$ be i.i.d. samples from a fixed but unknown distribution $\distr$ over the set of possible requests $\inputset$. For $\delta>0$, with probability at least $1-\delta$ we have
	\[
	\OPTDP -\REW_{\vgamma} \le  O\mleft(\frac{1}{\rho}\sqrt{T\log\mleft(mT/\delta\mright)}\mright) + \cump +\cumd,
	\]
	where $\REW_{\vgamma}\defeq\sum_{t}  f_t(\vx_t)$ is the reward of the algorithm for the sequence of inputs $\vgamma$.
\end{theorem}

The proof is based on two steps. First, by applying the Azuma-Hoeffding inequality, we show that, in the first $\tau\in[T]$ rounds, the average reward and cost for each resource $i$ up to $\tau$ is \emph{close}, with high probability, to $\E_{\vx\sim\bar\vxi}[\bar f(\vx)]$ and $\E_{\vx\sim\bar\vxi}[\bar c(\vx)[i]]$, where $\bar\vxi$ is the average of the strategy mixtures selected by the primal regret minimizer.
Then, we define a two-player, zero-sum, \emph{expected Lagrangian game} such that, by \cref{lemma:strong duality 2}, the value at the Nash equilibrium of game is equal to $\OPTLPE$. Finally, we show that $\bar\vxi$ is an approximation of the equilibrium strategy of one of the two players.

Analogously to the adversarial case (\cref{remark:regret high prob}), \cref{thm:stochastic} extends to the bandit feedback setting with minor modifications, whenever a primal regret minimizer for set $\cX$ with high-probability regret guarantees with respect to $\hat\lossp$ is available.

%% file: nonstationary.tex
\section{Regret Bound for the Non-Stationary Input Setting} \label{sec:nonstationary}

Here we consider the non-stationary input model from \citet{balseiro2022best}, where requests are drawn from independent but not necessarily identical distributions at each time step. This input model generalizes the stochastic setting of \cref{sec:stoc}. In particular, we show that our framework is robust to adversarial corruptions to i.i.d. stochastic inputs. 
This constitutes an interesting middle ground between the i.i.d. input model, which may be too optimistic, and the adversarial input model, which may be too pessimistic in practice. This intermediate input model captures problems such as malicious perturbations to the input (e.g., click fraud), and non-malicious perturbations (e.g., traffic spikes due to exogenous events). Other works studying adversarial corruptions to i.i.d. models are, for example, \cite{lykouris2018stochastic,chen2019robust}.

Given probability measures $\distr_1, \distr_2$, let $\|\distr_1 - \distr_2\|_{\text{TV}}$ be the total variation distance between $\distr_1$ and $\distr_2$. We will consider the set of sequences of probability measures with bounded mean total variation $\text{MD}(\vec{\distr}) = \sum_{t=1}^T \|\distr_t - \distr\|_{\text{TV}}$ around a reference distribution $\distr$:
\newcommand{\epsnonstat}{\ensuremath{\EuScript{E}_{\text{NS}}}}
\begin{align*}
    \mathcal C^{ID}(\epsnonstat) = \left\{ \vec{\distr} \in \Delta(\mathcal I)^T : \text{MD}(\vec{\distr}) \leq \epsnonstat \right\}
\end{align*}
Let $L^{\min} = \min_{\vxi, \vlambda, f,c} L(\vxi, \vlambda, f,c)$ and 
let $\Delta L = \max_{\vxi, \vlambda, f,c} L(\vxi, \vlambda, f,c) - L^{\min}$.
Now we define the \emph{underlying} Lagrangian game as the game played when using the reference distribution $\distr$, $\bar L(\vxi,\vlambda) = \mathbb E_{(f,c)\sim \distr}[L(\vxi,\vlambda,f,c)]$ and the round $t$ Lagrangian game as $\bar L_t(\vxi,\vlambda) = \mathbb E_{(f,c)\sim \distr_t}[L(\vxi,\vlambda,f,c)]$.

First we note the useful fact that the Lagrangian games at each time step are similar to the underlying Lagrangian game:
\begin{lemma}\label{lemma:nonstat similar expected game}
    For $\vec{\distr} \in \mathcal C^{ID}(\epsnonstat)$,
    the Lagrangian games satisfy the following two inequalities
    \[
        \left| \bar L_t(\vxi, \vlambda) - \bar L(\vxi, \vlambda) \right| \leq \Delta L \|\distr_t - \distr\|_{\text{TV}}, \forall t \in [T]
    \]
    \[
        \left| \frac{1}{\tau} \sum_{t=1}^\tau \bar L_t(\vxi, \vlambda) - \bar L(\vxi, \vlambda) \right|
        \leq \Delta L \epsnonstat/ \tau \quad \forall \tau \in [T].
    \]
\end{lemma}

Next we derive an analogue of \cref{lemma:expected game}.
\begin{lemma}\label{lemma:expected game nonstat}
Given $\tau\in[T]$, for $\delta\in(0,1)$, with probability at least $1-2\delta$, the average strategy mixture $\bar\vxi\in\Xi$ up to $\tau$ is such that, for any $\vlambda\in\cD$,   
\[
\bar L(\bar\vxi,\vlambda)
\ge  \OPTLPE- \frac{1}{\tau}\mleft(\cump+\cumd+4(1+1/\rho)\sqrt{2T\log(T/\delta)} - 2(1+1/\rho)\epsnonstat \mright) .
\]
\end{lemma}

Now we prove the main theorem. It is analogous to that of \cref{thm:stochastic}, but with the additional caveat that we must take into account the nonstationarity.

\begin{theorem}\label{thm:nonstat}
Consider Meta-Algorithm~\ref{alg:meta alg} equipped with two arbitrary regret minimizers $\cRp$ and $\cRd$ for the sets $\Xi$ and $\cD$, respectively. In particular, assume that they guarantee a cumulative regret up to time $T$ which is upper bounded by $\cump$ and $\cumd$, respectively. For each $t\in[T]$, let the inputs$(f_t,c_t)$ be i.i.d. samples from a fixed but unknown distribution $\distr_t$ over the set of possible requests $\inputset$ such that $\vec{\distr} \in \mathcal C^{ID}(\epsnonstat)$. For $\delta>0$, with probability at least $1-\delta$ we have
\[
\OPTDP -\REW_{\vgamma} \le  O\mleft(\frac{1}{\rho}\sqrt{T\log\mleft(mT/\delta\mright)}\mright) + \cump +\cumd + (3+3/\rho)\epsnonstat,
\]
where $\REW_{\vgamma}\defeq\sum_{t}  f_t(\vx_t)$ is the reward of the algorithm for the sequence of inputs $\vgamma$,
and $\OPTDP$ is the optimal dynamic policy under the reference distribution $\distr$.
\end{theorem}

\cref{thm:nonstat} shows that our algorithm can perform almost as well as the optimal dynamic policy (up to the non-stationarity error $\epsilon$) on an arbitrary reference distribution $\distr$. Specifically, we can let the reference distribution be the uniform mixture of the per-time-step distributions, $\distr = \frac{1}{T} \sum_{t=1}^T \distr_t$, and we get that if the mean deviation around the uniform mixture is small, then we get a strong performance guarantee.

%% file: appl.tex
\section{Applications}\label{sec:app}
In this section, we first provide an explicit instantiation of our algorithm in the classical multi-armed bandit with knapsack setting of \citet{badanidiyuru2013bandits}.
Then, we describe two applications of our framework to well known game-theoretic problems. This provide further evidence of the flexibility of our framework.

\input{bwk}
\input{stackelberg}

\input{firstPrice}

%% file: bwk.tex
\subsection{Multi-Armed Bandits with Knapsacks}

Consider a multi-armed bandit problem with $K$ arms (\emph{i.e.}, $\cX=[K]$), and per-arm utility and cost defined as in \cref{sec: preliminaries}. 

Let the primal regret minimizer for bandit feedback $\cRp$ be EXP3.P \citep{auer2002nonstochastic}. By its definition (see \cref{alg:meta alg}) and by \cref{eq:dual set}, the loss $\lossp$ is such that $\lossp:\Xi\to[-1/\rho,1]$. Then, at time $T$, EXP3.P guarantees that, with probability at least $1-\delta$, $\cump[T,\delta]\le O(\sqrt{KT\log(T/\delta)}/\rho )$.
\begin{corollary}
Consider Meta-Algorithm~\ref{alg:meta alg}, and let the primal regret minimizer for bandit feedback $\cRp$ be EXP3.P, and let the dual regret minimizer with full feedback $\cRd$ be online mirror descent with negative entropy reference function (see \cref{remark:omd}). 
We have the following two cases:
\begin{OneLiners}
\item If requests are chosen by an oblivious adversary. Letting $\alpha\defeq 1/\rho$, for each $\delta>0$ we have
\[
\OPTFD_{\vgamma} -\alpha\,\REW_{\vgamma}\le O(\alpha^2 \sqrt{KT\log(Tm/\delta)});
\]

\item if, for each $t\in[T]$, the inputs $(f_t,c_t)$ are i.i.d. samples from a fixed but unknown distribution $\distr$ over the set of possible requests $\inputset$. For $\delta>0$, we have
\[
\OPTDP -\REW_{\vgamma} \le  O\mleft(\sqrt{KT\log\mleft(Tm/\delta\mright)}/\rho\mright) ;
\]
\end{OneLiners}

with probability at least $1-\delta$, where $\REW_{\vgamma}\defeq\sum_{t}  f_t(\vx_t)$ is the reward of the algorithm for the sequence of inputs $\vgamma$.
\end{corollary}

%% file: stackelberg.tex
\subsection{Repeated Stackelberg Games with Knapsacks}

In Stackelberg games a \emph{leader} commits to a (possibly mixed) strategy, and then a \emph{follower} best responds to that strategy \citep{Stackelberg34:Marktform}.
Stackelberg games have recently received significant attention for their applications in security domains \citep{tambe2011security}. In such settings, Online \emph{Stackelberg Security Games} (SSG) have been introduced to circumvent the assumption that the leader must know the attacker’s utility function \citep{balcan2015commitment}.
We show that our framework could be extended to model repeated SSGs in which there are hard budget constraints with respect to deployed defensive resources.

We take the perspective of the leader that, at each time $t \in [T]$, plays a game against a follower of an unknown type.
The leader has a finite set of available actions $\cA_\LDR$ with $n_\LDR\defeq |\cA_\LDR|$, and strategies $\cX\defeq\Delta(\cA_\LDR)$, while the follower has a set of available actions $\cA_{\FLL}$ with $n_\FLL\defeq|\cA_{\FLL}|$, and strategies $\cY\defeq\Delta(\cA_{\FLL})$. 
The utility function of the follower at time $t$ is denoted by $u_t:\cX\times\cY\to\R$. We assume that, for each $t\in[T]$, $\vx\in\cX$, and $\vy\in\cY$, $u_t(\vx,\vy)\defeq \vx^\top\, U_t\, \vy$, for $U_t\in \R^{n_\LDR\times n_\FLL}$.
Moreover, we assume that the follower has a finite set of possible types $\cK$ and hence, for each $t$, $U_t\in\mleft\{U_k\mright\}_{k \in \cK}$.
At each $t$, the leader commits to a strategy $\vx_t\in\cX$. Then, the follower will play their \emph{best-response} given $\vx_t$.
Formally, for type $k\in\cK$, the follower plays the strategy $\vy^k_{\vx_t} \defeq \ve_a$, where $a\in\argmax_{a' \in \cA_\FLL}\vx_t^\top U_k \ve_{a'}$, and $\ve_a$ denotes the vector where component $a$ is equal to 1 and the others are equal to $0$.
As it is customary in the literature, we assume that the follower breaks ties in favour of the leader.
Then, the leader's utility function $f_t:\cX\to[0,1]$ is such that, for each $\vx\in\cX$, and $t$,
$
f_t(\vx)\defeq \vx^\top \, U_\LDR \, \vy^{k_t}_{\vx},
$
where $k_t$ is the follower type at round $t$. 
This function is upper semicontinuous, and it is therefore Borel measurable.

At each $t$, the leader pays a cost based on the strategy they commit to. In particular, for each $t$ there's a cost matrix $C_t\in[0,1]^{n_\LDR \times m}$, that specifies a vector of $m$ costs for leader's actions. The cost incurred by the leader at time $t$ is then $\vx_t^\top C_t$, where $\vx_t$ is the strategy played by the leader at time $t$.
The leader has an overall budget $B\in\R_{\ge0}$ for each resource. Let $\rho$ be the per-iteration budget (defined as in \cref{sec: preliminaries}),
Moreover, we assume the leader has a void action which yields no reward and no consumption of any other resource other than time. %

In order to apply \cref{alg:meta alg}, we show that there exists a regret minimizer for the leader. We show that this is possible despite the fact that the leader's utility is non-convex, and not even continuous. 
By \cref{remark:regret min x xi}, we can safely restrict our attention to regret minimizers that provide no-regret with respect to the optimal fixed strategy in $\vx^\ast\in \cX$.

As a first step, we show that for each sequence of follower's types $(k_t)_{t=1}^\tau$, there always exists an optimal mixed strategy belonging to a \emph{finite} set of strategies $\cX^\ast\subset \cX$. 
Moreover, we show that $\cX^\ast$ is independent from the sequence of types. In order to define the restricted set $\cX^\ast$, for each type $k \in \cK$ and action $a \in \cA_{\FLL}$, let $\cX^{k,a}\subseteq \cX$ be the set of leader's strategy in which $a$ is a best response for the follower of type $k$, \emph{i.e.}, $\cX^{k,a}\defeq \{\vx \in \cX:\vy^k_{\vx} = \ve_a\}$.
Let $\va\in\cA_{\LDR}^{|\cK|}$ be a tuple with an action per follower's type, and $\cX^{\va}$ be the polytope such that each action $\va[k]$ is optimal for the corresponding type $k$, \emph{i.e.}, $\cX^{\va} \defeq \cap_{k \in \cK} \cX^{k,\va[k]}$.
Finally, we let $\cX^\ast\defeq\cup_{\va \in \cA_{\LDR}^{|\cK|}} V(\cX^{\va})$, where $V(\cX^{\va})$ denotes the set of the vertexes of the polytope $\cX^{\va}$.

\begin{lemma}\label{lm:OptSmall}
	Let $\ell_{\LDR,t}(\vx,\vlambda)\defeq f_t(\vx) -  \langle\vlambda,\vx^{\top} C_t\rangle$ for all pairs $(\vx,\vlambda)$.
	Then, for each $\tau \in [T]$, each sequence of receiver's types $(k_t)_{t=1}^\tau$, and each sequence $(\vlambda_t)_{t = 1}^\tau$, it holds:
	\[
	\max_{\vx^\ast \in \cX^\ast} \sum_{t =1}^\tau  \ell_{\LDR,t}(\vx^\ast,\vlambda_t)  = 
	\max_{\vx \in \cX} \sum_{t =1}^\tau \ell_{\LDR,t}(\vx,\vlambda_t). 
	\]
\end{lemma}

Then, we bound the cardinality of the restricted set $\cX^\ast$.

\begin{lemma}\label{lemma:stackelberg 2}
	It holds $|\cX^*| \le (|\cK|n_{\FLL}^2)^{n_{\LDR}-1}$.
\end{lemma}

Lemma~\ref{lm:OptSmall} implies that to build a regret minimizer for $\cX$, it is sufficient to have small regret with respect to the optimal action in $\cX^\ast$.
Thus, we can focus on a regret minimizer for $\cX^\ast$.
Since $\cX^\ast$ has finite support, the set of randomized strategy $\Xi^\ast$ is the simplex over $\cX^\ast$, \emph{i.e.}, $\Xi^\ast=\Delta^{\cX^\ast}$.
As a primal regret minimizer, we can employ OMD with a negative entropy regularizer that provides regret upper bound $O( \sqrt{T\log(|\cX^\ast|)})$.
Therefore, we proved the existence of a regret minimizer for the primal decision space. 

\begin{theorem}
There exists a primal regret minimizer $\cRp$ for the Strackelberg problem with regret $\cump[T]=O(\sqrt{T n_{\LDR} \log(|\cK|n_{\FLL})} )$.
\end{theorem}

Equipped with the above result, we can directly apply \cref{thm:adversarial,thm:stochastic} to our setting.

%% file: firstPrice.tex
\subsection{Budget-Pacing in Repeated First-Price Auctions with Finite sets of Valuations and Bids}\label{sec:firstprice1}

Internet advertising platforms typically offer advertisers the possibility to
pace the rate at which their budget is depleted, through \emph{budget-pacing mechanisms} \citep{agarwal2014budget,conitzer2021multiplicative,balseiro2021budget}.
These mechanisms are essential to ensure that the advertisers' budget is not depleted too early (thereby missing potentially valuable future advertising opportunities), while being fully depleted within the planned duration of the campaign. 
We focus on budget pacing in the context of first-price auctions, which is particularly relevant for selling display ads (\emph{e.g.}, in 2019 Google announced a shift to first-price auctions for its AdManager exchange).\footnote{See \url{https://tinyurl.com/chv5nxys}.}
Dual mirror descent schemes which are usually employed in the context of repeated second price auctions \citep{balseiro2019learning,balseiro2022best} cannot be applied to our setting, because they rely on particular features of second price auctions.
In particular, in the primal update step, $\vx_t$ is chosen by maximizing a function of $f_t$ and $c_t$. Therefore, in general, this assumes that the decision maker observers $(f_t,c_t)$ before taking the decision at time $t$. The fact that costs are determined through a second price auction allows the decision maker to implicitly take the $\max$ without actually observing the costs, by bidding their \emph{adjusted} valuation. However, this is not possible when allocations are determined through first-price auctions.

We consider the problem faced by a bidder that takes part to a sequence of first-price auctions under budget constraints. In this section, we consider a simplified setting in which the bidder has a finite set of available valuation and bids. In \cref{sec:first price continuous} we show how to address the continuous case. 

At each round $t\in [T]$, the bidder observes their valuation $v_t$ extracted by a finite set of possible valuations $\cV\subset [0,1]$, with $\nval\defeq |\cV|$. In ad auctions this models the fact that the auctioneer shares with the advertisers some targeting information about users.
Then, the bidder chooses $b_t\in \cB$, where $ \cB \subset [0,1]$ is a finite set of $\nbid$ possible bids.
The utility of the bidder depends on the maximum among the competing bids, which we denote by $ m_t$.
In particular, if $b_t\ge m_t$, the bidder wins the auction, pays to the auctioneer $b_t$, and has utility $f_t(b_t)= v_t- b_t$. Moreover, the bidder incurs a cost $c_t(b_t)= b_t$. Otherwise, the bidder does not win the item, in which case $f_t(b_t)=0$, and $c_t(b_t)=0$.
Finally, the bidder has a budget $B \in \mathbb{R}_+$, which limits the total amount that the agent can spend throughout the $T$ rounds.
As a benchmark to evaluate the performance of the algorithm, we consider the best static policy $\pi:\cV\rightarrow \cB$. Next, we show that the problem can be easily addressed via our framework.
The set of static policies can be represented by $\cX\defeq \cB^{\nval}$, where a vector $\vb \in B^{\nval}$ is such that $\vb[v]$ is the bid played by the policy with valuation $v$.
Then, the utility function is such that $f_t(\vb)=(v_t-\vb[v_t]) \indicator{\vb[v_t]\ge m_t}$, where $I$ denotes the indicator function, and the cost is $c_t(\vb)= \vb[v_t] \indicator{\vb[v_t]\ge m_t}$. The set of strategy mixtures is given by the set $\Xi\defeq\Delta^{\cX}$.

Let $\lossb:\Xi\ni\vxi\mapsto\E_{\vb\sim \vxi}[f_t(\vb)]-\lambda_t \E_{\vb\sim \vxi}[c_t(\vb)]$ be the primal loss function. We want to show the existence of a regret minimizer for the set $\Xi$. To do that, by \cref{remark:regret min x xi}, we know that it is enough to design a regret minimizer for $\cX$. Then, by letting $\Xi^\ast\defeq(\Delta^{\cB})^{\nval}$, and since 
$
    \max_{\vb \in \cX} \sum_{t =1}^\tau f_t(\vb)-\lambda_t c_t(\vb)=\max_{\vxi \in \Xi^\ast} \sum_{t=1}^\tau \lossb(\vxi),
$
it is enough to design a regret minimizer for the set $\Xi^\ast$.
Since the primal loss function $\lossb$ is linear in $\vxi$, we can apply OMD with negative entropy regularizer to get a regret upper bound of $O(\sqrt{T \nval\log(\nbid)})$ (see, \emph{e.g.}, \citet{Farina21:Better}).
\begin{theorem}
    There exists a primal regret minimizer $\cRp$ for the problem of bidding in first-price auctions with regret upper bound $O(\sqrt{T\nval \log(\nbid)})$.
\end{theorem}
This immediately implies that \cref{thm:adversarial,thm:stochastic} hold in the full information setting (\emph{i.e.}, when the bidder observes $m_t$ for each $t$). In the bandit setting, one can obtain analogous results by instantiating an appropriate regret minimizer (\emph{e.g.}, \textsc{Exp3} by \citet{auer2002nonstochastic}) for each $v\in\cV$.

%% file: pacing.tex
\subsection{Budget Pacing in First-Price Auctions with Continuous Valuations and Bids}\label{sec:first price continuous}

We now generalize the simplified setting of \cref{sec:firstprice1} to the general case in which the bidder has an arbitrary set of possible valuations $\cV\subseteq [0,1]$ and bids $\cB\subseteq [0,1]$. Let %
$\ps$ be the set of static policies $\pi:\cV\to\cB$ mapping valuations to bids. 
Unlike the model considered by~\citet{badanidiyuru2021learning}, in which some structural assumptions on the inputs are made, we assume that valuation $v_t$ and $\mb_t$ can be arbitrarily chosen by an oblivious adversary (\emph{i.e.}, the adversary is agnostic to the private randomization used by the bidder).

\paragraph{Baseline} 

Let
\[
\psl\defeq \mleft\{ \p\in\ps: | \pi(v) - \pi(v') | \le \,|v-v'|,\,\forall v,v'\in\cV \mright\}
\]
be the collection of $1$-Lipschitz-continuous policies.  As pointed out by \citet{han2020learning}, a natural choice for the baseline in this setting is the best fixed feasible policy in $\psl$. This baseline allows the bidder to adapt their bid to their private valuation for the item, while only requiring the mild and natural constraint that the dependence of the bid on the private valuation is smooth (i.e., bids for items of similar value should be similar).

The set of strategy mixtures $\Xi$ is the set of probability measures over Borel sets of $\psl$, and the primal loss function is 
\[
\lossb:\Xi\ni\vxi\mapsto\E_{\p\sim \vxi}[f_t(\p(v_t))]-\lambda_t \E_{\p\sim \vxi}[c_t(\p(v_t))].
\]
We want to show the existence of a regret minimizer for the set $\Xi$. To do that, by \cref{remark:regret min x xi}, we know that it is enough to design a regret minimizer for $\psl$. Then, the cumulative external regret of the primal player is defined as the difference between the baseline $\sup_{\p^\ast\in\psl}\sum_{t=1}^T\mleft( f_t(\p^\ast(v_t))- \lambda_t c_t(\p^\ast(v_t))\mright)$, and the expected utility provided by the algorithm across the whole time horizon (a formal definition is provided in \cref{sec:first price alg}).

\subsubsection{Relevant Sets of Policies}

Let $\cB_{\epsilon}\defeq  \{0,\epsilon,\dots,1\}$ be a discretization of the set of possible bids with step $\epsilon$, and $\cV_{\epsilon}\defeq \{0,\epsilon,\dots,1\}$ be a discretization of the set of possible valuations with step $\epsilon$. Moreover, let $\disc{x}_y$ be the largest $z\in\R_{\ge0}$ such that $z\le x$, and $z$ is divisible by $y$, \emph{i.e.}, we discretize the value $x$ with step $y$.
 
As a first step, we show that for each $\epsilon\in[1/T,1]$, we can choose a \emph{finite} set of (deterministic) \emph{discretized policies} $\hat\ps_\epsilon$, that is guaranteed to include a policy with regret at most $2\epsilon T$ with respect to the optimal $\p^\ast\in\psl$. We define the set of policies $\hat\ps_\epsilon$ to be the subset of the finite set of policies $\hat\p:\cV_\epsilon\to\cB_{1/T}$ such that $|\hat \p(v)-\hat \p(v')|\le 2|v-v'|$ for each $v,v' \in \cV_\epsilon$. Finally, we define the set of \emph{thresholded policies}, with the goal of ensuring that, given a discretized policy,  the utility of the primal player is non-negative: given a valuation $v$ and a dual variable $\lambda$, thresholded policies ensure that, whenever the bid $b$ is greater than the highest competing bid (\emph{i.e.}, the bidder is winning the item), then the bid is such that $v-(1+\lambda)\,b\ge0$. A thresholded policy is built starting from a discretized policy $\hat\p\in\hat\ps_\epsilon$, for some $\epsilon\in[1/T,1]$. When the bidder follows $\hat\pi$ we apply an additive correction of $2\epsilon$, which, intuitively, corrects for underbidding due to discretization. 
\begin{definition}[Thresholded policy]\label{def:thresholded} 
	Given $\epsilon\in[1/T,1]$, let $\hat\p\in\hat\ps_\epsilon$ be a deterministic policy for the finite set $\cV_\epsilon$. Then, the \emph{thresholded policy} for $\hat\p$ is a function mapping elements of $\cV\times\cD$ into $\cB$ such that,
	\[
	\cV\times\cD\ni (v,\lambda)\mapsto \min\mleft\{\hat\p(\disc{v}_\epsilon) + 2\epsilon\,,\,\frac{v}{1+\lambda}\mright\}.
	\]
	Moreover, we denote by $\gamma$ the function that maps every  $\hat\p\in\hat\ps_\epsilon$ into the corresponding thresholded policy, and we let  $\Phi_\epsilon\defeq\mleft\{ \gamma(\hat\pi):\hat\p\in\hat\ps_\epsilon \mright\}$ be the set containing the thresholded policy of each policy in $\hat\ps_\epsilon$.
\end{definition}
An important property of thresholded policies is that they always provide positive utility in $[0,1]$. This is not the case for general policies which can have a negative utility $-1/\rho$.

We defined three sets of policies:
First and second, $\psl$ and  $\hat\ps_\epsilon$ map valuations (resp., discretized valuations) to bids (resp., discretized bids). The former, is the set of 1-Lipschitz-continuous policies mapping the original set of valuations $\cV$ into $\cB$. The latter is the set of 2-Lipschitz-continuous discretized policies, mapping $\cV_\epsilon$ into $\cB_{1/T}$. 
The third set of policies is the set of thresholded policies $\Phi_\epsilon$, which turns policies from $\hat\ps_\epsilon$ into mappings from $\cV\times \cD$ to $\cB$, where valuations and bids are \emph{not} discretized. Notice that thresholded policies can exploit more information than standard policies in $\ps$, because they can take into account the current value of the dual variable. 
The set of thresholded policies is particularly useful since, given any discretization of valuations with step $\epsilon$, there exists a policy in $\Phi_\epsilon$ which guarantees to the primal player utility \emph{close} to that of the baseline $\p^\ast$. The following result characterizes the relationship between the sets of policies $\psl$ and $\hat\ps_\epsilon$.

 \begin{lemma}\label{lemma:apxDiscretized}
 	For any sequence $(v_t,m_t,\lambda_t)_{t\in[T]}$ and $\epsilon\in[1/T,1]$, it holds
 	$$\sup_{\p^\ast\in\psl} \mleft\{\sum_{t \in [T]}\mleft( f_t(\p^\ast(v_t))- \lambda_t c_t(\p^\ast(v_t))\mright)\mright\}- \max_{\phi\in\Phi_\epsilon}\mleft\{ \sum_{t\in[T]}\mleft(f_t(\phi(v_t,\lambda_t))- \lambda_t c_t(\phi(v_t,\lambda_t))\mright)\mright\}\le \frac{7 \epsilon T}{\rho} .$$
 \end{lemma}

 \subsubsection{Hierarchical Tree Structure}\label{sec:tree structure}
 
The primal regret minimizer will be based on a hierarchical chaining of regret minimizers which will be organized in a tree structure. The procedure is similar to the one proposed by \citet{cesa2017algorithmic}, and later employed by \citet{han2020learning} to develop bidding algorithms for repeated first-price auctions in the unconstrained setting. In our setting, modifications are necessary in order to handle the dual player that manages the resource constraints. 
Hierarchical chaining provides two main advantages with respect to employing a single regret minimizer instantiated over the space of discretized policies. First, it reduces the number of experts  each regret minimizer can choose from. Second, in our chaining, the hierarchical tree structure is carefully designed so that each regret minimizer chooses among similar experts. 
This implies that each regret minimizer achieves a better regret bound than the one achievable in a classical prediction with expert advice problem. In turn, this implies a better asymptotic regret bound in terms of the dependence on $T$.

Let $\cH$ be the set of nodes of the tree. Each node is uniquely identified by its \emph{history} (i.e., by the sequence of edges on the path from the root to such node). We denote by $\cZ$ the set of terminal nodes of the tree, and by $\hroot\in\cH$ the root of the tree.
Edges of the tree represent policies. Given $h\in\cH\setminus\{\hroot\}$, we define $\parent{h}$ to be the last policy on the path from the root $\hroot$ to node $h$. 
Given an $m \in \mathbb{N}$, let $\epsilon_m\defeq2^{-m}$.
Then, given a sequence of $M\ge 2$ discretization levels $\epsilon_1,\epsilon_2,\ldots,\epsilon_M$, the tree is built level-by-level, by endowing nodes with appropriate edges which correspond to discretized policies. The granularity of the discretization increases as we move deeper in the tree. We define a collection of subsets $\cH_0,\cH_1,\ldots,\cH_M$ of the set of nodes $\cH$ as follows. Let $\cH_0\defeq \mleft\{\hroot\mright\}$. For $m\in[M]$, $\cH_m$ is the subset of nodes in the tree such that their parent policy $\parent{h}$ is a discretized policy in $\hat\ps_{\epsilon_m}$. Formally, for $m\in[M]$,
$\cH_m\defeq \mleft\{ h\in\cH: \parent{h}\in\hat\ps_{\epsilon_m} \mright\}$.

\begin{figure}[th]
 	\centering
 	\raisebox{1cm}{ \input{tree_example}}
 	\caption{In this simple example, the first level of the hierarchical structure (i.e., $\cH_1$) contains 3 nodes. Moreover, we observe that $|A(\hroot)|=4$, since $A(h)$ contains one edge for each of the child node in $\cH_1$, and one edge $\zchild{\hroot}$ which directly jumps to a terminal node belonging to $\cZ$. }
 	\label{fig:tree}
\end{figure}
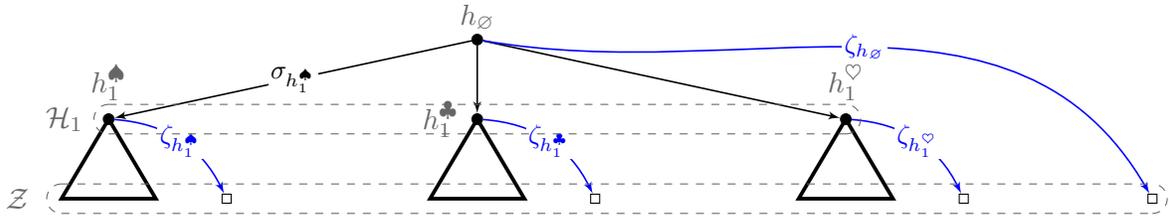

Given a node $h\in\cH_m$, with $m\in[M-1]$, we denote by $\cC(h)$ the set of policies with value discretization of step $\epsilon_{m+1}$ available at $h$. In particular, for each $h\in\cH_m$, $m\in[M-1]$,
\[
\cC(h)\defeq 	\mleft\{ \hat\p\in\hat\ps_{\epsilon_{m+1}}: \hat\p(v) = \parent{h}(v),\,\,\forall v\in\cV_{\epsilon_m} \mright\}.
\] 
Notice that $\cV_{\epsilon_m}\subset \cV_{\epsilon_{m+1}}$ by the definition of $\epsilon_m$.
For each level of the tree $m\in[M-1]$ and for each $h\in\cH_m$ we define a policy $\zchild{h}\in\hat\ps_{\epsilon_M}$ such that, for each $v\in\cV_{\epsilon_M}$ we have $\zchild{h}(v)\defeq \sigma_h(\disc{v}_{\epsilon_m}) + \disc{2(v-\disc{v}_{\epsilon_m})}_{1/T}$. Intuitively, this policy bids higher than any other policy in the subtree rooted in $h$.  
Now we can define the structure of the tree by specifying the set of policies (i.e., edges) available at each node. Given $h\in\cH$ the set of policies (edges) available at $h$ is defined as 
\[
A(h)\defeq \mleft\{\hspace{-1.25mm}\begin{array}{l}
	\displaystyle
	\cC(h)\cup\mleft\{\zchild{h}\mright\}\hspace{.5cm}\text{\normalfont if } h\in\cH_m,\, m\in[M-1]\\ [2mm]
	\emptyset \hspace{2.1cm}\text{\normalfont if } h\in\cZ\\
\end{array}\mright..
\] 
The above construction implies that nodes in $\cZ$ are either nodes at depth $M$ in the tree (i.e., $z$ is reachable through a sequence of $M$ policies),
or nodes $z$ resulting from ``jumps'' from intermediate levels of the tree $m\in[M-1]$ to the level of terminal nodes (i.e., $\sigma_z=\zchild{h}$ for some $h\in\cH_m$, $m\in[M-1]$). See \cref{fig:tree} for a visual depiction of this structure. 
For a node $h\in\cH\setminus\cZ$ and a policy $\pi\in A(h)$, we denote by $h\cdot\pi$ the node reached by following the edge corresponding to $\pi$ at $h$.

We observe that different edges in the tree may correspond to the same policy. In particular, a terminal node $z$ cannot be uniquely identified by the last policy played on the path from the root $\hroot$ to $z$ (but it can be identified by the sequence of edges on the path). In principle, we may have $z,z'\in\cZ$, with $z\ne z'$, such that $\parent{z} = \parent{z'}$.

 \subsubsection{Building a hierarchical policy}\label{sec:hierarchical policies}

 The final bidding strategy will consist of a mixture distribution over terminal nodes of the tree. When terminal node $z\in\cZ$ is selected according to this distribution, the bidder bids according to the policy $\parent{z}$ leading to the terminal node $z$. 
The set of policies available to the bidder is thus $\terminalp \defeq \mleft\{\sigma_z: z\in\cZ \mright\}$ (as already mentioned, it may occur that $|\terminalp|\le|\cZ|$). We now introduce the necessary notation to define how one such mixture distribution can be computed starting from a set of probability distributions specifying, for each node $h\in\cH\setminus\cZ$, the probability of selecting policies in $A(h)$ at $h$.  
 
 At each $h\in\cH\setminus\cZ$ and for each $t\in[T]$, $\vq_{t,h}\in\Delta^{A(h)}$ is such that $\vq_{t,h}[\p]$ is the probability of choosing the edge corresponding to policy $\p\in A(h)$ at $h$.
 Moreover, for $t\in[T]$, and given a distribution $\vq_{t,h'}$ for each $h'\in\cH\setminus\cZ$, we denote by $\vp_{t,h}\in\Delta^{\terminalp}$ the
 vector specifying the probability of selecting each policy in $\terminalp$ when starting from a given node $h\in\cH$. For any $h$, $\vp_{t,h}$ can be computed from local distributions $\vq_{t,\cdot}$ in a bottom-up fashion, by starting from terminal nodes of the tree. In particular, for $h\in\cH$ and $\pi\in\terminalp$, it is enough to apply the following recursive definition 
\begin{equation}\label{eq:p distr}
 \vp_{t,h}[\pi]\defeq \mleft\{\hspace{-1.25mm}\begin{array}{l}
 	\displaystyle
 	\sum_{\pi'\in A(h)} \vq_{t,h}[\pi']\,\vp_{t,h\cdot\pi}[\pi]\hspace{.5cm}\text{\normalfont if } h\in\cH_m,\, m\in[M-1]\\ [2mm]
 	\indicator{\pi=\parent{h}} \hspace{2.6cm}\text{\normalfont if } h\in\cZ\\
 \end{array}\mright..
\end{equation}
 
 In the following section we show how to update local distributions $\vq_{t,h}$ for each $h\in\cH\setminus\cZ$ so that the resulting mixture of policies at the root of the tree $\vp_{t,\hroot}$ can be used to obtain a good approximation of the value attained with the best fixed policy in $\psl$.

 \subsubsection{Chaining Algorithm}\label{sec:first price alg}

We are now ready to present a hierarchical chaining algorithm which can serve as a primal regret minimizer for the ``continuous'' first-price setting which we are considering. The algorithm exploits the hierarchical tree structure described in \cref{sec:tree structure}, and it maintains, for each node $h\in\cH\setminus\cZ$ of the tree and time $t\in[T]$, a \emph{local} policy mixture $\vq_{t,h}$, and a mixture $\vp_{t,h}$ over policies in $\terminalp$, as described in \cref{sec:hierarchical policies}.
At each $t$, the algorithm updates in a bottom-up manner the strategy $\vq_{t,h}$ and the mixture $\vp_{t,h}$ for each node $h$. The former is updated proportional to the exponential weights of the cumulative rewards of each policy available at the current node; the latter is updated by computing the distribution induced over policies in $\terminalp$ by strategies in lower levels of the tree, together with the newly computed local strategy $\vq_{t,h}$.
After the updates, the bidder selects a policy according to $\pi_t\sim\vp_{t,\hroot}$ and places a bid following the thresholded  policy $\gamma(\pi_t)$. 
Note that thresholded policies require knowing the value of the current dual variable in order to make a decision (see \cref{def:thresholded}). Thus, since our primal regret minimizer will use thresholded policies,  it needs to observe the dual variable before making a decision. This is not a problem since we can let the regret minimizers play sequentially, as we observe in Remark~\ref{remark:sequential}.

\Cref{alg:first price} provides a detailed description of the primal regret minimizer for the first-price setting. The algorithm receives as input the time horizon $T$, the number of levels $M$ of the hierarchical tree structure (see \cref{sec:tree structure}), one discretization step $\epsilon_m\defeq2^{-m}$ per level of the tree $m\in[M]$, and the per-iteration budget $\rho$. At each $t\in [T]$, the algorithm observes the valuation $v_t$ and the dual variable $\vlambda_t$.
Then, the algorithm proceeds level-by-level, starting from level $M-1$ up to the root of the tree. For each level $m$ and each node $h\in\cH_m$, the algorithm computes a suitable learning rate $\eta_{t,h}\gets\min\{1/4,\sqrt{log(|A(h)|)/(t\,\Delta_m\,\epsilon_{m+1}  })\}$, where $\Delta_m\defeq 4(1+1/\rho)\epsilon_m$. Then, this learning rate is used to update the strategy $\vq_{t,h}$ using an exponential weights update in \cref{eq: alg exp}. Mixtures over $\terminalp$ are updated according to \cref{eq: alg p}. We observe that $\vp_{t,h}$ for $h\in\cZ$ are initialized consistently with \cref{eq:p distr}. After having updated $\vq_{t,\hroot}$ and $\vp_{t,\hroot}$ (i.e., after reaching the last node in the sequence of updates, which is the root), the algorithm draws a bidding policy $\pi_t$ from $\terminalp$ according to $\vp_{t,\hroot}$. Then, this policy is mapped to the corresponding thresholded policy $\gamma(\pi_t)$, which is used so select a bid as a function of the current valuation $v_t$ and dual multiplier $\lambda_t$. After observing the highest competing bid $m_t$, the algorithm computes the instantaneous reward at time $t$ for each policy $\pi\in\terminalp$ as follows
\[
r(\pi,\lambda_t,v_t,m_t)\defeq \indicator{b_t\ge m_t}\,\mleft(v_t-(1+\lambda_t)b_t\mright),
\]
where $b_t\defeq \gamma(\pi)(v_t,\lambda_t)$ is the bid specified by the thresholded policy $\gamma(\pi)$ for valuation $v_t$ and dual variable $\lambda_t$. 
When clear from the context, given $t\in[T]$ and policy $\pi$, we will write $r_t(\pi)$ in place of $r(\pi,\lambda_t,v_t,m_t)$. 
Then, given a node $h\in\cH_m$ with $m\in[M-1]$, and a policy $\pi\in A(h)$, the expected reward at time $t$ is computed as $\E_{\vp_{t,h\cdot\pi}}[r(\pi',\lambda_t,v_t,m_t)]$, where $\pi'\in\terminalp$ is drawn according to the mixture $\vp_{t,h\cdot \pi}$ of the node reached from $h$ by selecting $\pi$. The algorithm updates cumulative rewards for each node $h$ and policy in $A(h)$ accordingly, as specified in \cref{eq: alg rew}.

\begin{algorithm}[th] 
	\caption{Primal regret minimizer for ``continuous'' first-price setting.}
	\label{alg:first price}
	\KwData{time horizon $T$, number of levels of the tree $M\defeq \lfloor \log_2\sqrt{T}\rfloor$, brackets $\epsilon_m\defeq 2^{-m}$ for $m\in [M]$, budget per iteration $\rho$, $\Delta_m\defeq 4(1+1/\rho)\epsilon_m$ for each $m\in[M]$.}
	
	\textbf{Inizialization: } $\vr_{0,h}[\pi]\gets 0$ for all non-terminal node $h\in\cH\setminus \cZ$ and $\pi\in A(h)$.
	
	\For{$t = 1, 2, \ldots , T$}{
	Observe valuation $v_t\in[0,1]$ and dual variable $\vlambda_t$\\
	Set $\vp_{t,h}[\pi]\gets \indicator{\pi=\parent{h}}$  for each $(h,\pi)\in\cZ\times\terminalp$\\
	\For{$m=M-1,M-2,\ldots,0$}{
		\For{$h\in\cH_m$}{
				Set learning rate $\eta_{t,h}\gets\min\mleft\{1/4,\sqrt{log(|A(h)|/(t\,\Delta_m\,\epsilon_{m+1}  })\mright\}$\\
				For each $\pi\in A(h)$
				\begin{equation}\label{eq: alg exp}
					\vq_{t,h}[\pi]\gets \frac{\exp\mleft(\eta_{t,h}\,\vr_{t-1,h}[\pi]\mright)}{\sum_{\pi'\in A(h)}\exp\mleft( \eta_{t,h}\,\vr_{t-1,h}[\pi'] \mright)}
				\end{equation}

				For each $\pi\in\terminalp$
				\begin{equation}\label{eq: alg p}
					\vp_{t,h}[\pi]\gets \sum_{\pi'\in A(h)} \vq_{t,h}[\pi']\,\vp_{t,h\cdot\pi}[\pi]
				\end{equation} 
			}
		}
	Draw $\pi_t\sim\vp_{t,\hroot}$ and play according to $\gamma(\pi_t)$\\
	Observe the highest competing bid $m_t$\\
	\For{$m=M-1,M-2,\ldots,0$}{
		\For{$h\in\cH_m$ and $\pi\in A(h)$}{
			\begin{equation}\label{eq: alg rew}
				\vr_{t,h}[\pi]\gets \vr_{t-1,h}[\pi] + \sum_{\pi'\in\terminalp} \vp_{t,h\cdot\pi}[\pi']\, r(\pi',\lambda_t,v_t,m_t).
			\end{equation}
			}
		}
	}
\end{algorithm}

In order to prove that \Cref{alg:first price} yields a regret minimizer for the primal problem, we need to provide a suitable regret upper bound for 
\[
\regp[T]\defeq\sup_{\p^\ast\in\psl}\sum_{t=1}^T\mleft( f_t(\p^\ast(v_t))- \lambda_t c_t(\p^\ast(v_t)) - \E_{\pi\sim\vp_{t,\hroot}}\mleft[r_t(\pi)\mright]\mright).
\]
We start by defining a notion of a \emph{good edge} as proposed by \citet{han2020learning}. Intuitively, we say that an edge (i.e., a policy) is \emph{good} if it has near-optimal instantaneous reward at each time $t\in[T]$.
Given a policy $\pi\in\hat\ps_\epsilon$, a valuation $v\in[0,1]$ and a dual variable $\lambda\in\cD$, in order to increase readability in the remainder of the section we write $\pi(v,\lambda)$ to denote $\gamma(\pi)(v,\lambda)$ (i.e., the value of the thresholded policy corresponding to the discrete policy $\pi$). Then, we can formally define the notion of \emph{good edge} as follows.

\begin{definition}[Good edge]\label{def:goodExpert}
	For each level $m \in [M-1]$ and node $h \in \cH_{m}$, an edge $\pi \in A(h)$ is $\Delta$-good, with $\Delta\in[0,1]$, if, for all $\pi' \in A(h) $ and $t \in [T]$, it holds
	\[
	\E_{\pi'' \sim \vp_{t,h\cdot\pi}} \mleft[r(\pi'',\lambda_t,v_t,m_t)\mright] \ge  \E_{\p'' \sim \vp_{t,h\cdot\pi'}}\mleft[r(\pi'',\lambda_t,v_t,m_t)\mright]-\Delta.
	\]
\end{definition}

We now show that or each $m \in [M-1]$, and node $h \in \cH_m$, the edge $ \zchild{h}$ is a good edge. 
The intuition for this fact is that every edge other than $\zchild{h}$ at $h$ is, in a sense, a refinement of $\zchild{h}$: due to the Lipschitzness of the policies and the fact that all policies in the subtree rooted at $h$ must agree with $\zchild{h}$ on the discretization $\mathcal{V}_{\epsilon_m}$, each policy cannot improve over $\zchild{h}$ by too much.
In particular, the policy $\zchild{h}$ bids higher than any policy in the subtree rooted at $h$, and hence wins the item whenever a policy in the subtree win the item. At the same time, it never bids much higher than the policies in the subtree by Lipschitzness.

\begin{lemma}\label{lemma:delta good}
	For each $m \in [M-1]$ and each $h \in \cH_m$, the edge $\zchild{h}$ is $\Delta_m$-good for $\Delta_m\defeq 4(1+1/\rho)\epsilon_m$.
\end{lemma}

Equipped with the result on the existence of $\Delta_m$-good policies (\cref{lemma:delta good}), we can now upper bound the regret accumulated at each node of the tree. In particular, this can be done by exploiting a result provided by \citet[Theorem 3]{han2020learning}. 
Let 
\[
R_T^h\defeq \max_{\hat\pi\in A(h)}\sum_{t\in[T]}\mleft( \E_{ \pi\sim \vp_{t,h_{m}\cdot\hat\pi}}\mleft[r_t(\pi)\mright]- \E_{\pi\sim \vq_{t,h_{m}}}\E_{ \pi'\sim \vp_{t,h_{m}\cdot\pi}}\mleft[r_t(\pi')\mright] \mright).
\]
be the ``local'' cumulative regret accumulated at node $h\in\cH\setminus\cZ$.

\begin{lemma}\label{lemma:regret good expert}
Given $m \in [M]$ and a node $h \in \mathcal{H}_m$, the exponential-weights policy with a time-varying learning rate $\eta_t = \min\{1/4, \sqrt{\log |A(h)|/(t\Delta_m)}\}$ guarantees the following regret upper bound: 
	\[
	R_{T}^h \le 4\sqrt{T\Delta_m\log |A(h)|)} + 32(4+\log T)\log |A(h)|. 
	\]
\end{lemma}

Finally, by exploiting the above result, we can now prove the following regret upper bound for \cref{alg:first price}.

\begin{theorem} \label{thm:firstPrice}
	Algorithm~\ref{alg:first price} guarantees regret
	    \begin{align*}
	        \regp[T]\le  O\mleft( \frac{1}{\rho}\sqrt{T} \log^2(T)\mright).
	    \end{align*}  
\end{theorem}

These are, to the best of our knowledge, the first theoretical guarantees for budget-pacing in repeated first-price auctions without structural assumptions on the inputs. However, the primal regret minimizer that we employ (\cref{alg:first price}) is not efficient: 
it needs to instantiate a regret minimizer for each of the exponentially-many nodes $\cH\setminus\cZ$ of the tree. 
Even if the computational complexity of the algorithm is a limitation for practical applications, we believe that this is an important first step in designing efficient budget-pacing algorithms for first-price auctions. 
Our results show that, in order to provide an efficient algorithm, it is sufficient to design an efficient primal regret minimizer, and then such an algorithm can be plugged into our framework. 
Therefore, future advances in no-regret algorithms for the first-price setting can be applied to the case with budgets by plugging those algorithms into our framework. 
For example, \citet{han2020learning} provide an inefficient no-regret algorithm for first-price auctions without budgets, similar to the one we provide here, and show how to modify it to obtain an efficient algorithm. 
In the future, a similar approach could be used to develop an efficient primal regret minimizer for our setting.
Since the focus of this is on providing a general framework for online learning problems with resource constraints, we leave this as an interesting future research direction.

We conclude the section by stating our final results for budget-pacing in first-price auctions, which we obtain from \cref{thm:adversarial} and \cref{thm:stochastic} by instantiating the framework with the primal regret minimizer described in \cref{alg:first price}.

\begin{corollary}
Consider Meta-Algorithm~\ref{alg:meta alg}, and let the primal regret minimizer $\cRp$ be \cref{alg:first price}, and the dual regret minimizer $\cRd$ be online mirror descent with negative entropy reference function (see \cref{remark:omd}). 
We have the following two cases:
\begin{OneLiners}
\item If requests are chosen by an oblivious adversary. Letting $\alpha\defeq 1/\rho$, for each $\delta>0$ we have
\[
\OPTFD_{\vgamma} -\alpha\,\REW_{\vgamma}\le O(\alpha^2 \sqrt{T\log^2(Tm/\delta)});
\]

\item if, for each $t\in[T]$, the inputs $(f_t,c_t)$ are i.i.d. samples from a fixed but unknown distribution $\distr$ over the set of possible requests $\inputset$. For $\delta>0$, we have
\[
\OPTDP -\REW_{\vgamma} \le  O\mleft(\sqrt{T\log^2\mleft(Tm/\delta\mright)}/\rho\mright) ;
\]
\end{OneLiners}

with probability at least $1-\delta$, where $\REW_{\vgamma}\defeq\sum_{t}  f_t(\vx_t)$ is the reward of the algorithm for the sequence of inputs $\vgamma$.
\end{corollary}

%% file: tree_example.tex
\begin{tikzpicture}[>=latex',baseline=0pt,scale=.98]
	\def\done{5}
	\def\dtwo{.40*1.6}
	\def\dleaf{.22*1.6}
	\def\dvert{-.9*1.2}
	
	\node[fill=black,draw=black,circle,inner sep=.5mm] (A) at (0, 0) {};

	\node[fill=black,draw=black,circle,inner sep=.5mm] (X) at ($(-\done,\dvert)$) {};
	\node[fill=black,draw=black,circle,inner sep=.5mm] (Z) at ($(\done,\dvert)$) {};
	\node[fill=black,draw=black,circle,inner sep=.5mm] (Y) at ($(0,\dvert)$) {};

	\draw[black,line width=0.5mm] ($(X) + (-\dtwo, \dvert)$) -- ($(X) + (\dtwo, \dvert)$) -- ($(X)$) -- cycle;
	\node[fill=white,draw=black,inner sep=.6mm] (t1) at ($(X)+(2.5*\dtwo,\dvert)$) {};
	
	\draw[black,line width=0.5mm] ($(Y) + (-\dtwo, \dvert)$) -- ($(Y) + (\dtwo, \dvert)$) -- ($(Y)$) -- cycle;
	\node[fill=white,draw=black,inner sep=.6mm] (t2) at ($(Y)+(2.5*\dtwo,\dvert)$) {};
	
	\draw[black,line width=0.5mm] ($(Z) + (-\dtwo, \dvert)$) -- ($(Z) + (\dtwo, \dvert)$) -- ($(Z)$) -- cycle;
	\node[fill=white,draw=black,inner sep=.6mm] (t3) at ($(Z)+(2.5*\dtwo,\dvert)$) {};
	
	\node[fill=white,draw=black,inner sep=.6mm] (t4) at ($(t3)+(4*\dtwo,0)$) {};

	\draw[semithick] (A) edge[->] node[fill=white,inner sep=.9] {\small{$\parent{h^\spadesuit_1}$}} (X);
	\draw[->,semithick] (A) --(Y);
	\draw[->,semithick] (A) --(Z);
	
	\node[black!60!white]  at ($(A) + (0, .3)$) {$\hroot$};
	\node[black!60!white]  at ($(X) + (0, .5)$) {$h^\spadesuit_1$};
	\node[black!60!white]  at ($(Y) + (-.5,0)$) {$h^\clubsuit_1$};
	\node[black!60!white]  at ($(Z) + (0,.5)$) {$h^\heartsuit_1$};
	
	\draw[semithick] (X) edge[->,bend left,blue] node[fill=white,inner sep=.9] {\small{$\zchild{h^\spadesuit_1}$}} (t1);
	\draw[semithick] (Y) edge[->,bend left,blue] node[fill=white,inner sep=.9] {\small{$\zchild{h^\clubsuit_1}$}} (t2);
	\draw[semithick] (Z) edge[->,bend left,blue] node[fill=white,inner sep=.9] {\small{$\zchild{h^\heartsuit_1}$}} (t3);
	\draw[semithick] (A) edge[->,out=-10,in=120,blue] node[fill=white,inner sep=.9] {\small{$\zchild{\hroot}$}} (t4);

	\draw[black!60!white,dashed] ($(X) + (0, .2)$) arc (90:270:.2);
	\draw[black!60!white,dashed] ($(X) + (0, .2)$) -- ($(Z) + (0, .2)$);
	\draw[black!60!white,dashed] ($(X) + (0, -.2)$) -- ($(Z) + (0, -.2)$);
	\draw[black!60!white,dashed] ($(Z) + (0, -.2)$) arc (-90:90:.2);
	\node[black!60!white,dashed]  at ($(X) + (-.6, 0)$) {$\cH_1$};
	
	\draw[black!60!white,dashed] ($(X) + (-\dtwo, \dvert) + (0, .2)$) arc (90:270:.2);
	\draw[black!60!white,dashed] ($(X) + (-\dtwo, \dvert) + (0, .2)$) -- ($(t4) + (0, .2)$);
	\draw[black!60!white,dashed] ($(X) + (-\dtwo, \dvert) + (0, -.2)$) -- ($(t4) + (0, -.2)$);
	\draw[black!60!white,dashed] ($(t4) + (0, -.2)$) arc (-90:90:.2);
	\node[black!60!white]  at ($(X) + (-\dtwo, \dvert) + (-.6, 0)$) {$\cZ$};

\end{tikzpicture}

%% file: tools.tex
\clearpage
\appendix
\section{Omitted Proofs}\label{sec:omitted proofs}

\subsection{Proofs for \cref{sec: mixtures}}

\noindent{\bf Theorem~\ref{strong duality}.}
{\em	Let $f:\cX\to[0,1]$, $c:\cX\to[0,1]^m$, $(f,c)\in\cI$. It holds:
\[
\sup_{\vxi\in\Xi }\inf_{\vlambda\ge 0} L(\vxi,\vlambda,f, c)=\inf_{\vlambda\ge 0}\sup_{\vxi\in\Xi } L(\vxi,\vlambda, f, c)=\OPTLP_{f,c}.
\]}
\begin{proof}{Proof.}
\xhdr{Auxiliary sets.} Let 
\[
\cV\defeq \mleft\{ \mleft(\vv,t\mright)\in\R^{m+1}: \exists \vxi\in\Xi \,\,\text{\normalfont s.t. }\, \E_{\vx\sim\vxi}\mleft[ c(\vx)[i]\mright]-\rho\le \vv[i],\forall i\in[m],\, 
\E_{\vx\sim\vxi}\mleft[f(\vx)\mright]\ge t\mright\},
\]
\[
\cW\defeq \mleft\{(\vec{0},w)\in\R^{m+1}:\, w> \OPTLP_{f,c}\mright\},
\]
where each element of $\cW$ is composed of an $m$-dimensional vector of zeros, and a scalar $w$.
Notice that the dimension of the two sets does not depend on the dimensionality of $\Xi$. In particular, $\cV,\cW$ have finite dimension even when we have an infinite-dimensional space of strategy mixtures.
We claim that $\cV$ and $\cW$ are convex, and $\cV\cap\cW=\varnothing$. Take any two points $(\vv_1,t_1)\in\cV$, $(\vv_2,t_2)\in\cV$, and $\alpha\in[0,1]$. Then, let $\vxi_\alpha=\alpha\,\vxi_1+(1-\alpha)\,\vxi_2$, where $\vxi_1$ (resp., $\vxi_2$) is a point in $\Xi$ for which the constraints of $\cV$ for $(\vv_1,t_1)$ (resp., $(\vv_2,t_2)$) are satisfied. 
We have that $\vxi_\alpha\in\Xi$. Moreover, by linearity of expectation, we have, for each resource $i\in[m]$, $\E_{\vx\sim\vxi_\alpha}\mleft[ c(\vx)[i]\mright]-\rho\le \alpha \vv_1[i]+(1-\alpha)\vv_2[i]$, and $\E_{\vx\sim\vxi_\alpha}\mleft[ f(\vx)\mright]\ge \alpha t_1+(1-\alpha)t_2$. Then, $\alpha (\vv_1,t_1) + (1-\alpha)\,(\vv_2,t_2)\in \cV$ (i.e., $\cV$ is convex). 
It is immediate to check that $\cW$ is convex. 
Finally, assume that there exists a point $(\vv',t')\in \cV\cap\cW$. By definition of $\cV$, there exists $\vxi'\in\Xi$ such that $\E_{\vx\sim\vxi'}\mleft[ c(\vx)\mright]-\vrho\le \vv'$, and $\E_{\vx\sim\vxi'}\mleft[ f(\vx)\mright]\ge t'$. Then, by definition of $\cW$, $\vv'=\vec{0}$, that is, $\vxi'$ is budget-feasible. However, the fact that $\E_{\vx\sim\vxi'}\mleft[ f(\vx)\mright]>\OPTLP_{f,c}$ is in contradiction with $\OPTLP_{f,c}$ being the optimal value of a feasible solution to LP \eqref{eq:opt lp gen}.

\xhdr{Separating $\cV$ and $\cW$.} By assumption we have that the primal objective $\OPTLP_{f,c}$ is finite (otherwise, if $\OPTLP_{f,c}=+\infty$, we could immediately recover our result by weak duality).
The sets $\cV$ and $\cW$ are convex and do not intersect. Therefore, by the separating hyperplane theorem, there exists a point $(\tilde\vlambda,\tilde\mu)\in\R^{m+1}\setminus\{\vec{0}\}$, and a scalar value $\beta\in\R$ such that 
\begin{equation}\label{eq: sep hyperplane 1}
(\vv,t)\in\cV\,\,\implies\,\, \langle\tilde\vlambda ,v\rangle + \tilde\mu \,t \le \beta,
\end{equation}
\begin{equation}\label{eq: sep hyperplane 2}
(\vec{0},w)\in\cW\,\,\implies\,\, \tilde\mu\, w \ge \beta.
\end{equation}
First, it must be that $\tilde\vlambda[i]\le 0$ for all $i\in[m]$, and $\tilde\mu\ge 0$, otherwise we would get unboundedness of the lefthand side of \cref{eq: sep hyperplane 1}. 
Moreover, since \cref{eq: sep hyperplane 2} must hold for each $w>\OPTLP_{f,c}$, by continuity we have $\tilde\mu\,\OPTLP_{f,c}\ge\beta$. For each $\vxi\in\Xi$ there exists a pair $(\vv_{\vxi},t_{\vxi})\in\cV$ such that $\E_{\vx\sim\vxi}\mleft[ c(\vx)\mright]-\vrho= \vv_{\vxi}$, and $\E_{\vx\sim\vxi}\mleft[ f(\vx)\mright]=t_{\vxi}$. Together with the fact that \cref{eq: sep hyperplane 1} holds for each $(\vv,t)\in\cV$, this yields that for each $\vxi\in\Xi$,
\begin{equation}\label{eq: sep 3}
\langle \tilde\vlambda, \E_{\vx\sim\vxi}\mleft[ c(\vx)\mright]-\vrho\rangle+\tilde\mu\,\E_{\vx\sim\vxi}\mleft[ f(\vx)\mright]\le \beta\le \tilde\mu\,\OPTLP_{f,c}.
\end{equation}
If $\tilde\mu>0$, then for each $\vxi\in\Xi$
\[
\frac{1}{\tilde\mu}\langle \tilde\vlambda, \E_{\vx\sim\vxi}\mleft[ c(\vx)\mright]-\vrho\rangle+\E_{\vx\sim\vxi}\mleft[ f(\vx)\mright]\le \OPTLP_{f,c}.
\]
By letting $\hat\vlambda=-\tilde\vlambda / \tilde\mu$, we have $L(\vxi,\hat\vlambda,f,c)\le \OPTLP_{f,c}$ for each $\vxi$. In particular, $\sup_{\vxi} L(\vxi,\hat\vlambda,f,c)\le\OPTLP_{f,c}$. Then, 
\[\inf_{\vlambda\ge 0}\,\sup_{\vxi\in\Xi} L(\vxi,\vlambda,f,c)\le\OPTLP_{f,c}.\] 
By weak duality we get $\inf_{\vlambda\ge 0}\sup_{\vxi} L(\vxi,\vlambda)=\OPTLP_{f,c}$, which proves our statement.

\noindent If $\tilde\mu=0$, from \cref{eq: sep 3} we get that, for each $\vxi\in\Xi$, $\langle\tilde\vlambda, \E_{\vx\sim\vxi}\mleft[ c(\vx)\mright]-\vrho\rangle\le 0$. Let $\vxi_\nullx\defeq\delta_\nullx \in\Xi$ be the Dirac mass that plays the null action.  
We have that $\E_{\vx\sim\vxi_{\nullx}}\mleft[ c(\vx)\mright]-\vrho< 0$. Then, since $\tilde\vlambda\le 0$, it must be $\tilde\vlambda=0$. That is in contradiction with $(\tilde\vlambda,\tilde\mu)\ne\vec{0}$.
This concludes the proof.
\end{proof}

\bigskip
\noindent{ \bf\  Lemma~\ref{lemma:strong duality 2}.}
{\em Let $\cD$ be defined as in \cref{eq:dual set}. Given $f:\cX\to[0,1]$, $c:\cX\to[0,1]^m$, $(f,c)\in\cI$, it holds
\[
\hspace{-.2cm}
\sup_{\vxi\in\Xi }\inf_{\vlambda\in\cD} L(\vxi,\vlambda, f, c)=\inf_{\vlambda\in\cD}\sup_{\vxi\in\Xi } L(\vxi,\vlambda, f, c)=\OPTLP_{ f, c}.
\]}
\begin{proof}{Proof.}
As a first step, we show that 
\begin{subequations}
\begin{align*}
\inf_{ \vlambda\in\cD}\,\sup_{\vxi\in\Xi } L(\vxi,\vlambda,f,c)\le \inf_{\vlambda\ge 0}\,\sup_{\vxi\in\Xi} L(\vxi,\vlambda,f,c).
\end{align*}
\end{subequations}
To do so, notice that for any $\vlambda'$ with $||\vlambda'||_1>1/\rho$, we have \[\sup_{\vxi\in\Xi } L(\vxi,\vlambda',f,c) > 1 \ge \OPTLP_{f,c}= \inf_{\vlambda\ge 0}\,\sup_{\vxi\in\Xi} L(\vxi,\vlambda,f,c),\]
where the first inequality holds because the null action provides value at least $\langle\vlambda',\vrho \rangle >1$.
Then, we have:
\begin{align*}
\sup_{\vxi\in\Xi }\,\inf_{\vlambda\in\cD} L(\vxi,\vlambda,f,c) &\ge \sup_{\vxi\in\Xi }\,\inf_{\vlambda\ge 0} L(\vxi,\vlambda,f,c)\\
&=\OPTLP_{f,c}\\
&=\inf_{\vlambda\ge 0}\,\sup_{\vxi\in\Xi } L(\vxi,\vlambda,f,c)\\
&\ge \inf_{ \vlambda\in\cD}\,\sup_{\vxi\in\Xi} L(\vxi,\vlambda,f,c),
\end{align*}
where the first inequality holds since on the lefthand side we have a more restrictive set of dual variables, and the second and the third inequalities hold by strong duality.
Finally, by the  max–min inequality 
\[
\sup_{\vxi\in\Xi }\,\inf_{\vlambda\in\cD} L(\vxi,\vlambda,f,c)\le \inf_{ \vlambda\in\cD}\,\sup_{\vxi\in\Xi } L(\vxi,\vlambda,f,c).
\]
This proves our statement.
\end{proof}

\bigskip
\noindent{ \bf\  Lemma~\ref{lemma: stoc opt ub}.}
{\em Given a distribution over inputs $\distr$, let $\bar f:\cX\to[0,1]$ be the expected reward function, and $\bar c:\cX\to[0,1]^m$ be the expected resource-consumption function. Then, $T \cdot \OPTLPE\ge\OPTDP$.}

\begin{proof}{Proof.}
Let $\vlambda^\ast = (\lambda_1^\ast,\dots, \lambda^\ast_m)\in\cD$ be an optimal dual vector for LP \eqref{eq:opt lp gen} with functions $\bar f,\bar c$.
By strong duality (\cref{lemma:strong duality 2}), it holds 
\begin{equation} \label{eq:dualForAll}
\bar f(\vx)+\langle\vlambda^\ast, \vrho-\bar c(\vx)\rangle \le \OPTLPE,\,\, \forall \vx \in \cX.
\end{equation}
Moreover, let $(\vxi_t)_{t=1}^T$ be the sequence of strategy mixtures specified by a given policy $\psi$, and denote by $\vx_t\sim\vxi_t$ an action realization sampled according to the strategy mixture at $t$. Let 
\[
Z_t\defeq(T-t)\OPTLPE + \sum_{t' \in [t]}\mleft( f_{t'}(\vx_{t'})+\langle\vlambda ,\vrho- c_{t'}(\vx_{t'})\rangle\mright).\]
Then, by \cref{eq:dualForAll}, and since 
\[
    \E_{\vx \sim \vxi_t}\mleft[\bar f(\vx)+\langle\vlambda, \vrho-\bar c(\vx)\rangle\mright]=
    \E_{\vx \sim \vxi_t,\,(f_t,c_t)\sim \distr}\mleft[ f_t(\vx)+\langle\vlambda, \vrho-c_t(\vx)\rangle\mright],
\]
the stochastic process $Z_0,\dots,Z_T$ is a supermartingale. 
Let $\tau\in[T]$ be the stopping time of the algorithm, that is, when the algorithm depletes the first resource. Then, the realized utility is such that $\sum_{t=1}^\tau f_t(\vx_t)\le Z_\tau$.
This holds since 
\begin{align*}
(T-\tau)\OPTLPE + \sum_{t=1}^\tau\langle\vlambda, \vrho- c_{t}(\vx_t)\rangle &
\ge (T-\tau)\OPTLPE + \sum_{i=1}^m\vlambda[i] (\tau \rho- T\rho)
\\&
\ge (T-\tau) \mleft(\OPTLPE - \langle\vrho,\vlambda\rangle\mright) \ge 0,
\end{align*}
where the first inequality comes from the fact that $\sum_{t \in [\tau]} c_{t}(\vx_t)[i] \le B$ for each $i\in [m]$, and the last inequality comes from \cref{eq:dualForAll} with $\vx$ equal to the void action $\nullx$.
Then, taking the expectation on both sides, we get $\E[Z_{\tau}] \ge \E[\sum_{t=1}^\tau f_t(\vx_t)]$. Let $v_\psi$ be the value obtained through policy $\psi$. 
By Doob's optional stopping theorem, $T \cdot \OPTLPE= Z_0 \ge \E[Z_{\tau}]\ge v_\psi$. Then, since this holds for every possible policy $\psi$, we have $T \cdot \OPTLPE \ge \sup_{\psi \in \Psi} \E[v_\psi] = \OPTDP $. This concludes the proof.
\end{proof}

\subsection{Proofs for \cref{sec:adv}} \label{sec:appAdv}

In this section, we provide more details on why \cref{remark:regret high prob} holds. 
Consider a regret minimizer for bandit feedback guaranteeing with probability at least $1-\delta$ that
\begin{equation}\label{eq:reg bandit}
\hat\regp\defeq \sup_{\vx\in\cX} \sum_{t=1}^\tau \mleft(\hat\lossp(\vx)-\hat\lossp(\vx_t)\mright)\le \cump[T,\delta].
\end{equation}
Then, let
\[
w_t\defeq \hat\lossp(\vx_t)-\lossp(\vxi_t)=\hat\lossp(\vx_t)-\E_{\vx\sim\vxi_t}\mleft[\hat\lossp(\vx)\mright],
\]
and observe that $|w_t|\le 1+1/\rho$. By applying the Azuma-Hoeffding inequality we get that given a $\tau \in [T]$
\[
\Pr\mleft(\sum_{t=1}^\tau w_t> \underbrace{(1+1/\rho)\sqrt{2T\ln(1/\delta)}}_{\eqqcolon q(\delta)}\mright)\le \delta.
\]
Then, by a standard application of the union bound,
\[
\Pr\mleft(\forall \tau\in[T], \quad \sum_{t=1}^\tau w_t \le q(\delta) \mright)\le 1-T\cdot\delta.
\]
Then, with probability at least $1-\delta$,
\[
\sum_{t=1}^\tau\lossp(\vx_t)\le q(T/\delta) + \sum_{t=1}^\tau \E_{\vx\sim\vxi_t}\mleft[\hat\lossp(\vx)\mright].
\]
Then, by \cref{remark:regret min x xi} and by \cref{eq:reg bandit}, we obtain that, with probability at least $1-2\delta$, for each $\tau \in [T]$ 
\begin{align*}
    \sup_{\vxi\in\Xi}\sum_{t=1}^\tau\mleft(\lossp(\vxi)-\lossp(\vxi_t)\mright) & =\sup_{\vx\in\cX}\sum_{t=1}^\tau\mleft(\hat\lossp(\vx)-\lossp(\vxi_t)\mright)\\
    & \le \sup_{\vx\in\cX}\sum_{t=1}^\tau\mleft(\hat\lossp(\vx)-\lossp(\vx_t)\mright)+q(T/\delta)\\
    & \le \cump[T,\delta]+q(T/\delta).
\end{align*}
This show that \cref{remark:regret high prob} is verified.

\subsection{Proofs for \cref{sec:stoc}}

\noindent{ \bf\  Lemma~\ref{lemma:expected game}.}
{\em		Given $\tau\in[T]$, for $\delta\in(0,1)$, with probability at least $1-2\delta$, the average strategy mixture $\bar\vxi\in\Xi$ up to $\tau$ is such that, for any $\vlambda\in\cD$,   
		$
		\bar L(\bar\vxi,\vlambda)\ge  \OPTLPE- \frac{1}{\tau}\mleft(\cump+\cumd+4(1+1/\rho)\sqrt{2T\log(T/\delta)}\mright).
		$}
\begin{proof}{Proof.}
We proceed in two steps:
\paragraph{(1)} 
Let $\tau\in[T]$, and $\vxi^\ast$ be the optimal strategy up to $\tau$ for the primal player in hindsight (\emph{i.e.}, with knowledge of the sequence of $\vlambda_t$ up to $\tau$, while taking expectations over $f_t$ and $c_t$ at each $t$). Moreover, let $v^\ast$ be the minimax value for the expected Lagrangian game $(\Xi,\cD,\bar L)$.
Then,

\begin{align*}
    \frac{1}{\tau}\sum_{t=1}^\tau L(\vxi_t,\vlambda_t,f_t,c_t) & \ge \frac{1}{\tau}\sum_{t=1}^\tau L(\vxi^\ast,\vlambda_t,f_t,c_t) - \frac{1}{\tau}\cump[T]\\
    & \gewhp \frac{1}{\tau}\sum_{t=1}^\tau \bar L(\vxi^\ast,\vlambda_t) -\frac{1}{\tau}\mleft(\cump[T]+2(1+1/\rho)\sqrt{2T\log(T/\delta)}\mright)\\
    & = \sup_{\vxi}\bar L(\vxi,\bar \vlambda) -\frac{1}{\tau}\mleft(\cump[T]+2(1+1/\rho)\sqrt{2T\log(T/\delta)}\mright) \\
    & \ge \inf_{\vlambda}\sup_{\vxi}\bar L(\vxi,\vlambda) -\frac{1}{\tau}\mleft(\cump[T]+2(1+1/\rho)\sqrt{2T\log(T/\delta)}\mright)\\
    & = v^\ast -\frac{1}{\tau}\mleft(\cump[T]+2(1+1/\rho)\sqrt{2T\log(T/\delta)}\mright),
\end{align*}
where $\gewhp$ denotes statements that hold with probability at least $1-\delta$ and $\bar \vlambda\defeq \frac{1}{\tau}\sum_{t=1}^\tau \vlambda_t$.

\paragraph{(2)} Fix $\vlambda\in\cD$. We have,
\begin{align*}
    \frac{1}{\tau}\sum_{t=1}^\tau  L(\vxi_t,\vlambda_t,f_t,c_t) & \le \frac{1}{\tau}\sum_{t=1}^\tau  L(\vxi_t,\vlambda,f_t,c_t) + \frac{1}{\tau}\cumd[T]\\
    & \lewhp \frac{1}{\tau}\sum_{t=1}^\tau  \bar L(\vxi_t,\vlambda) + \frac{1}{\tau}\mleft(\cumd[T]+2(1+1/\rho)\sqrt{2T\log(T/\delta)}\mright)\\
    & = \bar L(\bar\vxi,\vlambda) + \frac{1}{\tau}\mleft(\cumd[T]+2(1+1/\rho)\sqrt{2T\log(T/\delta)}\mright).
\end{align*}
By \cref{lemma:strong duality 2}, we have $v^\ast=\OPTLPE$. Then, by combining the inequalities from Step (1) and (2), and by taking a union bound we get the result.
\end{proof}

\noindent{ \bf\  Theorem~\ref{thm:stochastic}.}
{\em Consider Meta-Algorithm~\ref{alg:meta alg} equipped with two arbitrary regret minimizers $\cRp$ and $\cRd$ for the sets $\Xi$ and $\cD$, respectively. In particular, assume that they guarantee a cumulative regret up to time $T$ which is upper bounded by $\cump$ and $\cumd$, respectively. For each $t\in[T]$, let the inputs$(f_t,c_t)$ be i.i.d. samples from a fixed but unknown distribution $\distr$ over the set of possible requests $\inputset$. For $\delta>0$, with probability at least $1-\delta$ we have
	\[
	\OPTDP -\REW_{\vgamma} \le  O\mleft(\frac{1}{\rho}\sqrt{T\log\mleft(mT/\delta\mright)}\mright) + \cump +\cumd,
	\]
	where $\REW_{\vgamma}\defeq\sum_{t}  f_t(\vx_t)$ is the reward of the algorithm for the sequence of inputs $\vgamma$.
}
\begin{proof}{Proof.}
Let $\tau\in[T]$ be the stopping time of \cref{alg:meta alg}, and $\bar\vxi\in\Xi$ be such that, for all $\vx\in\cX$, $\bar\vxi(\vx)\defeq\sum_{t=1}^\tau\vxi_t(\vx)/\tau$. The proof proceeds in two steps.

\xhdr{Step 1.} Consider the first $\tau\in[T]$ rounds. By applying the Azuma-Hoeffding inequality we have that the average reward and cost for each resource $i$ up to $\tau$ is \emph{close}, with high probability, to $\E_{\vx\sim\bar\vxi}[\bar f(\vx)]$ and $\E_{\vx\sim\bar\vxi}[\bar c(\vx)[i]]$, respectively. Formally, given $\tau \in [T]$, by letting $\err^0_{\tau,\delta}\defeq O(\sqrt{\tau\log(m/\delta)})$ with probability at least $1-\delta$, we have
\begin{align}\label{eq:stoc step 1 1}
& \frac{1}{\tau}\sum_{t=1}^\tau f_t(\vx_t)\ge \E_{\vx\sim\bar\vxi}\mleft[\bar f(\vx)\mright] - \frac{1}{\tau}\err^0_{\tau,\delta}\\ 
& \frac{1}{\tau}\sum_{t=1}^\tau c_t(\vx_t)[i] \le \E_{\vx\sim\bar\vxi}\mleft[\bar c(\vx)[i]\mright] + \frac{1}{\tau}\err^0_{\tau,\delta}\hspace{.2cm}\forall i\in[m].\label{eq:stoc step 1 2}
\end{align}

\xhdr{Step 2.} Consider a sequence of repeated two-player, zero-sum games up to a given time $\tau\in[T]$, in which Player 1 (i.e., the \emph{primal player}) chooses as their action $\vxi\in\Xi$, and Player 2 (i.e., the \emph{dual player}) chooses $\vlambda\in\cD$. For each $t\in[\tau]$, for a pair of actions $(\vxi,\vlambda)$, Player 1 (resp., Player 2) observes utility function $L(\vxi,\vlambda,f_t,c_t)$ (resp., $-L(\vxi,\vlambda,f_t,c_t)$). 
When $(f_t,c_t)$ are drawn i.i.d. from some fixed distribution $\distr$, we can define $\bar L(\vxi,\vlambda)\defeq \E_{(f,c)\sim\distr}\mleft[L(\vxi,\vlambda,f,c)\mright]$. We say that $(\Xi,\cD,\bar L)$ is the \emph{expected Lagrangian game}. %

Let us condition on the fact that \cref{eq:stoc step 1 1}, \cref{eq:stoc step 1 2}, and \cref{lemma:expected game} hold for each $\tau\in[T]$. This event holds with probability $1-3\delta T$ (by a standard application of the union bound). Then, let $\tau\in[T]$ be the stopping time of the algorithm. By definition of stopping time, there exists a resource $\stopi\in[m]$ such that $\sum_{t=1}^\tau c_t(\vx_t)[\stopi]>B-1$ (see the primal decision in \cref{alg:meta alg}). By taking $\hat\vlambda$ such that $\hat\vlambda[\stopi]=1/\rho$, and $\hat\vlambda[i]=0$ for $i\ne\stopi$, and by \cref{eq:stoc step 1 1,eq:stoc step 1 2}, we have
\begin{align*}
    \bar L\mleft(\bar\vxi,\hat\vlambda\mright) & = \E_{\vx\sim\bar\vxi}\mleft[\bar f(\vx)\mright] + \langle\hat\vlambda,\vrho-\E_{\vx\sim\bar\vxi}\mleft[\bar c(\vx)\mright]\rangle\\
    & \le  \frac{1}{\tau}\mleft(\sum_{t=1}^\tau f_t(\vx_t) + \tau - \frac{1}{\rho}\sum_{t=1}^\tau c_t(\vx_t)[\stopi]+(1+\frac{1}{\rho})\err^0_{\tau,\delta} \mright)\\
    & \le  \frac{1}{\tau}\mleft(\sum_{t=1}^\tau f_t(\vx_t) + \tau - T+ \frac{1}{\rho} +(1+\frac{1}{\rho})\err^0_{\tau,\delta}\mright).
\end{align*}
Then, plugging the above expression in \cref{lemma:expected game} yields the following
\begin{align*}
    \REW_{\vgamma} & =  \sum_{t=1}^\tau f_t(\vx_t)
    \ge \tau\,\OPTLPE + T - \tau - \frac{1}{\rho} 
    - \mleft( (1+\frac{1}{\rho})\err^0_{\tau,\delta} +\cump+\cumd+4(1+1/\rho)\sqrt{2T\log(T/\delta)}\mright).
\end{align*}
Then, by \cref{lemma: stoc opt ub}, and since $\OPTLPE\le 1$,
\begin{align*}
    \OPTDP - \REW_{\vgamma} &\le T\,\OPTLPE - \tau\,\OPTLPE - T + \tau + \frac{1}{\rho}
     + (1+\frac{1}{\rho})\err^0_{\tau,\delta} +\cump+\cumd+4(1+1/\rho)\sqrt{2T\log(T/\delta)}\\
    & \le \cump +\cumd + O\mleft(1/\rho\sqrt{T\log(mT/\delta)}\mright).
\end{align*}
Finally, to obtain a bound that holds with probability $\delta'>0$, it is sufficient to take $\delta=\frac{\delta'}{3T}$. This does not affect the bound except for a constant factor hidden in the $O\mleft(\cdot\mright)$ notation.
This concludes the proof.
\end{proof}

\subsection{Proofs for \cref{sec:nonstationary}}

\noindent{ \bf\  Lemma~\ref{lemma:nonstat similar expected game}} {\em
    For $\vec{\distr} \in \mathcal C^{ID}(\epsnonstat)$,
    the Lagrangian games satisfy the following two inequalities
    \[
        \left| \bar L_t(\vxi, \vlambda) - \bar L(\vxi, \vlambda) \right| \leq \Delta L \|\distr_t - \distr\|_{\text{TV}}, \forall t \in [T]
    \]
    \[
        \left| \frac{1}{\tau} \sum_{t=1}^\tau \bar L_t(\vxi, \vlambda) - \bar L(\vxi, \vlambda) \right|
        \leq \Delta L \epsnonstat/ \tau \quad \forall \tau \in [T].
    \]}
\begin{proof}{Proof.}
    Fix some $t\in [T]$.
    First we bound the difference $\bar L_t(\vxi, \vlambda) - \bar L(\vxi, \vlambda)$, which is equal to 
    \begin{align*}
        \bar L_t(\vxi, \vlambda) - \bar L(\vxi, \vlambda)
        & = \int_{(f,c)} L(\vxi,\vlambda,f,c) d\distr_t - \int_{(f,c)} L(\vxi,\vlambda,f,c) d\distr \\
        & = \int_{(f,c)} (L(\vxi,\vlambda,f,c) - L^{\min}) d\distr_t - \int_{(f,c)} (L(\vxi,\vlambda,f,c) - L^{\min}) d\distr \\
        & \leq \int_{(f,c)} (L(\vxi,\vlambda,f,c) - L^{\min}) d (\distr_t - \distr)^+ \\
        & \leq \Delta L \int_{(f,c)}  d (\distr_t - \distr)^+ \\
        & \leq \Delta L \|\distr_t - \distr\|_{\text{TV}}
    \end{align*}
    Here $(\distr_t - \distr)^+$ is the positive part of the signed measure $(\distr_t - \distr)$, and the first inequality follows because all coefficients $(L(\vxi,\vlambda,f,c) - L^{\min})$ in the integral are nonnegative.
    The other direction can be shown analogously.
    This shows the first inequality of the lemma.
    It now follows that the second inequality of the lemma holds, by taking a sum over $t$ and invoking the definition of the mean total variation.
\end{proof}

\noindent{ \bf\  Lemma~\ref{lemma:expected game nonstat}}
{\em Given $\tau\in[T]$, for $\delta\in(0,1)$, with probability at least $1-2\delta$, the average strategy mixture $\bar\vxi\in\Xi$ up to $\tau$ is such that, for any $\vlambda\in\cD$,   
$
\bar L(\bar\vxi,\vlambda)
\ge  \OPTLPE- \frac{1}{\tau}\mleft(\cump+\cumd+4(1+1/\rho)\sqrt{2T\log(T/\delta)} - 2(1+1/\rho)\epsnonstat \mright) .
$
}
\begin{proof}{Proof.}
We proceed in two steps:
\paragraph{(1)} Let $\tau\in[T]$, and $\vxi^\ast$ be the optimal strategy up to $\tau$ for the primal player in hindsight. Moreover, let $v^\ast$ be the minimax value for the expected Lagrangian game $(\Xi,\cD,\bar L)$.
Then,
\begin{align*}
    \frac{1}{\tau}\sum_{t=1}^\tau L(\vxi_t,\vlambda_t,f_t,c_t) & \ge \frac{1}{\tau}\sum_{t=1}^\tau L(\vxi^\ast,\vlambda_t,f_t,c_t) - \frac{1}{\tau}\cump[T]\\
    & \gewhp \frac{1}{\tau}\sum_{t=1}^\tau \bar L_t(\vxi^\ast,\vlambda_t) -\frac{1}{\tau}\mleft(\cump[T]+2(1+1/\rho)\sqrt{2T\log(T/\delta)}\mright)\\
    & \geq \frac{1}{\tau}\sum_{t=1}^\tau \bar L(\vxi^\ast,\vlambda_t) -\frac{1}{\tau}\mleft(\cump[T]+2(1+1/\rho)\sqrt{2T\log(T/\delta)} + \Delta L \epsnonstat \mright) \\
    & = \sup_{\vxi}\bar L(\vxi,\bar \vlambda) -\frac{1}{\tau}\mleft(\cump[T]+2(1+1/\rho)\sqrt{2T\log(T/\delta)} + \Delta L \epsnonstat \mright) \\
    & \ge \inf_{\vlambda}\sup_{\vxi}\bar L(\vxi,\vlambda) -\frac{1}{\tau}\mleft(\cump[T]+2(1+1/\rho)\sqrt{2T\log(T/\delta)} + \Delta L \epsnonstat \mright)\\
    & = v^\ast -\frac{1}{\tau}\mleft(\cump[T]+2(1+1/\rho)\sqrt{2T\log(T/\delta)} + \Delta L \epsnonstat \mright),
\end{align*}
where $\gewhp$ denotes statements that hold with probability at least $1-\delta$. 
The first step follows by the definition of regret.
The second step follows, with high probability, by applying the Azuma-Hoeffding inequality to the martingale $X_t = X_{t-1} + \bar L_t(\vxi^\ast,\vlambda_t) - L(\vxi^\ast,\vlambda_t, f_t,c_t)$, here $L(\vxi^\ast,\vlambda_t, f_t,c_t)$ should be viewed as a random variable over $f_t,c_t$.
The third step follows by applying \cref{lemma:nonstat similar expected game}.
The fourth step follows from linearity in $\vlambda$, where $\bar \vlambda\defeq \frac{1}{\tau}\sum_{t=1}^\tau \vlambda_t$.

\paragraph{(2)} Fix $\vlambda\in\cD$. We have,
\begin{align*}
    \frac{1}{\tau}\sum_{t=1}^\tau  L(\vxi_t,\vlambda_t,f_t,c_t) & \le \frac{1}{\tau}\sum_{t=1}^\tau  L(\vxi_t,\vlambda,f_t,c_t) + \frac{1}{\tau}\cumd[T]\\
    & \lewhp \frac{1}{\tau}\sum_{t=1}^\tau  \bar L_t(\vxi_t,\vlambda) + \frac{1}{\tau}\mleft(\cumd[T]+2(1+1/\rho)\sqrt{2T\log(T/\delta)}\mright)\\
    & \le \frac{1}{\tau}\sum_{t=1}^\tau  \bar L(\vxi_t,\vlambda) + \frac{1}{\tau}\mleft(\cumd[T]+2(1+1/\rho)\sqrt{2T\log(T/\delta)} + \Delta L \epsnonstat \mright)\\
    & = \bar L(\bar\vxi,\vlambda) + \frac{1}{\tau}\mleft(\cumd[T]+2(1+1/\rho)\sqrt{2T\log(T/\delta)} + \Delta L \epsnonstat \mright).
\end{align*}
The first step follows by the definition of regret.
The second step by applying Azuma-Hoeffding using the same martingale as above.
The third step follows by applying \cref{lemma:nonstat similar expected game}.
The fourth step follows from linearity in $\vxi$.

By \cref{lemma:strong duality 2}, we have $v^\ast=\OPTLPE$. Then, by combining the inequalities from Step (1) and (2), using $\Delta L = 1+1/\rho$, and by taking a union bound we get the result.
\end{proof}

\noindent{ \bf\  Theorem~\ref{thm:nonstat}}
{\em Consider Meta-Algorithm~\ref{alg:meta alg} equipped with two arbitrary regret minimizers $\cRp$ and $\cRd$ for the sets $\Xi$ and $\cD$, respectively. In particular, assume that they guarantee a cumulative regret up to time $T$ which is upper bounded by $\cump$ and $\cumd$, respectively. For each $t\in[T]$, let the inputs$(f_t,c_t)$ be i.i.d. samples from a fixed but unknown distribution $\distr_t$ over the set of possible requests $\inputset$ such that $\vec{\distr} \in \mathcal C^{ID}(\epsnonstat)$. For $\delta>0$, with probability at least $1-\delta$ we have
\[
\OPTDP -\REW_{\vgamma} \le  O\mleft(\frac{1}{\rho}\sqrt{T\log\mleft(mT/\delta\mright)}\mright) + \cump +\cumd + (3+3/\rho)\epsnonstat,
\]
where $\REW_{\vgamma}\defeq\sum_{t}  f_t(\vx_t)$ is the reward of the algorithm for the sequence of inputs $\vgamma$,
and $\OPTDP$ is the optimal dynamic policy under the reference distribution $\distr$.
}

\begin{proof}{Proof.}
Let $\tau\in[T]$ be the stopping time of \cref{alg:meta alg}, and $\bar\vxi\in\Xi$ be such that, for all $\vx\in\cX$, $\bar\vxi(\vx)\defeq\sum_{t=1}^\tau\vxi_t(\vx)/\tau$. 
Moreover, let $\bar f, \bar c$ be the expected payoff and cost functions with respect to the \emph{underlying} distribution $\distr$.
The proof proceeds in two steps.

\paragraph{Step 1} Consider the first $\tau\in[T]$ rounds. By applying the Azuma-Hoeffding inequality we have that the average reward and cost for each resource $i$ up to $\tau$ is not too much worse, with high probability, than $\E_{\vx\sim\bar\vxi}[\bar f(\vx)] + \frac{1}{\tau}\epsnonstat$ and $\E_{\vx\sim\bar\vxi}[\bar c(\vx)[i]] + \frac{1}{\tau}\epsnonstat$, respectively. 
To see this,
first we define the martingale sequence $X_t = X_{t-1} + \mathbb E_{x\sim \vxi_t, f_t,c_t} f_t(x) - f_t(x_t)$, where $f_t(x_t)$ should be interpreted as a random variable over $f_t$ and $x_t$.
Given $\tau \in [T]$, by letting $\err^0_{\tau,\delta}\defeq O(\sqrt{\tau\log(m/\delta)})$, we have that with probability at least $1-\delta$, 
\begin{align}
 \frac{1}{\tau}\sum_{t=1}^\tau f_t(\vx_t)
&\ge \frac{1}{\tau} \sum_{t=1}^\tau \E_{\vx\sim\vxi_t, (f_t,c_t)\sim \distr_t} \mleft[ f_t(\vx)\mright] - \frac{1}{\tau}\err^0_{\tau,\delta}\nonumber\\ 
&\ge \frac{1}{\tau} \sum_{t=1}^\tau \E_{\vx\sim\vxi_t, (f_t,c_t)\sim \distr} \mleft[ f_t(\vx)\mright] - \frac{1}{\tau}\err^0_{\tau,\delta} - \frac{1}{\tau} \sum_{t=1}^\tau\| \distr_t - \distr\|_{\text{TV}}\nonumber\\ 
&\ge \frac{1}{\tau} \sum_{t=1}^\tau \E_{\vx\sim\vxi_t, (f_t,c_t)\sim \distr} \mleft[ f_t(\vx)\mright] - \frac{1}{\tau}\err^0_{\tau,\delta} - \frac{1}{\tau} \epsnonstat\nonumber\\
&= \E_{\vx\sim\bar\vxi} \mleft[ \bar f(\vx)\mright] - \frac{1}{\tau}\err^0_{\tau,\delta} - \frac{1}{\tau} \epsnonstat, \label{eq:nonstat step 1 1}
\end{align}

and by a similar argument
\begin{align}
& \frac{1}{\tau}\sum_{t=1}^\tau c_t(\vx_t)[i] 
\le \E_{\vx\sim\bar\vxi}\mleft[\bar c(\vx)[i]\mright] + \frac{1}{\tau}\err^0_{\tau,\delta} + \frac{1}{\tau} \epsnonstat\hspace{.2cm}\forall i\in[m]. \label{eq:nonstat step 1 2}
\end{align}

\paragraph{Step 2} 
Now we consider the underlying Lagrangian game $\bar L(\vxi,\vlambda)$ played  under the distribution $\distr$.
Let us condition on the fact that \cref{eq:nonstat step 1 1}, \cref{eq:nonstat step 1 2}, and \cref{lemma:expected game nonstat} hold for each $\tau\in[T]$. This event holds with probability $1-3\delta T$ (by a standard application of the union bound). Then, let $\tau\in[T]$ be the stopping time of the algorithm. By definition of stopping time, there exists a resource $\stopi\in[m]$ such that $\sum_{t=1}^\tau c_t(\vx_t)[\stopi]>B-1$ (see the primal decision in \cref{alg:meta alg}). By taking $\hat\vlambda$ such that $\hat\vlambda[\stopi]=1/\rho$, and $\hat\vlambda[i]=0$ for $i\ne\stopi$, and by \cref{eq:nonstat step 1 1,eq:nonstat step 1 2}, we have
\begin{align*}
    \bar L\mleft(\bar\vxi,\hat\vlambda\mright) & = \E_{\vx\sim\bar\vxi}\mleft[\bar f(\vx)\mright] + \langle\hat\vlambda,\vrho-\E_{\vx\sim\bar\vxi}\mleft[\bar c(\vx)\mright]\rangle\\
    & \le  \frac{1}{\tau}\mleft(\sum_{t=1}^\tau f_t(\vx_t) + \tau - \frac{1}{\rho}\sum_{t=1}^\tau c_t(\vx_t)[\stopi]+ (1+1/\rho) \left(\err^0_{\tau,\delta}+ \epsnonstat \right) \mright)\\
    & \le  \frac{1}{\tau}\mleft(\sum_{t=1}^\tau f_t(\vx_t) + \tau - T+ \frac{1}{\rho} + (1+1/\rho) \left(\err^0_{\tau,\delta}+ \epsnonstat \right) \mright).
\end{align*}
The last inequality follows because $(1/\rho)\sum_{t=1}^\tau c_t(\vx_t)[\stopi]>(1/\rho)(B-1) = T+1/\rho$.
Then, simplifying and plugging the above expression into \cref{lemma:expected game nonstat} yields the following
\begin{align*}
    \REW_{\vgamma}  = \sum_{t=1}^\tau f_t(\vx_t)
    &\ge \tau \bar L\mleft(\bar\vxi,\hat\vlambda\mright) -\tau + T - \frac{1}{\rho} - (1+1/\rho) \left(\err^0_{\tau,\delta}+ \epsnonstat \right) \\
    &\ge \tau\,\OPTLPE  -\tau + T - \frac{1}{\rho} 
    - \mleft( \cump+\cumd + (1+1/\rho) \left(\err^0_{\tau,\delta}+ 3\epsnonstat + 4\sqrt{2T\log(T/\delta)}\right)  \mright).
\end{align*}
Then, by \cref{lemma: stoc opt ub}, and since $\OPTLPE\le 1$,
\begin{align*}
    \OPTDP - \REW_{\vgamma} &\le T\,\OPTLPE - \tau\,\OPTLPE + \tau - T  + \frac{1}{\rho}
     + \cump+\cumd + (1+1/\rho) \left(\err^0_{\tau,\delta}+ 3\epsnonstat + 4\sqrt{2T\log(T/\delta)}\right) \\
    & \le \cump +\cumd + (3+3/\rho)\epsnonstat + O\mleft(1/\rho\sqrt{T\log(mT/\delta)}\mright).
\end{align*}
Finally, to obtain a bound that holds with probability $\delta'>0$, it is sufficient to take $\delta=\frac{\delta'}{3T}$. This does not affect the bound except for a constant factor hidden in the $O\mleft(\cdot\mright)$ notation.
This concludes the proof.
\end{proof}

\subsection{Proofs for \cref{sec:app}}

\noindent{ \bf\  Theorem~\ref{lm:OptSmall}.}
{\em	Let $\ell_{\LDR,t}(\vx,\vlambda)\defeq f_t(\vx) -  \langle\vlambda,\vx^{\top} C_t\rangle$ for all pairs $(\vx,\vlambda)$.
	Then, for each $\tau \in [T]$, each sequence of receiver's types $(k_t)_{t=1}^\tau$, and each sequence $(\vlambda_t)_{t = 1}^\tau$, it holds:
	\[
	\max_{\vx^\ast \in \cX^\ast} \sum_{t =1}^\tau  \ell_{\LDR,t}(\vx^\ast,\vlambda_t)  = 
	\max_{\vx \in \cX} \sum_{t =1}^\tau \ell_{\LDR,t}(\vx,\vlambda_t). 
	\]}
\begin{proof}{Proof.}
We show that given an optimal strategy $\vx \in \cX$, we can build a strategy $\vx^\ast \in \cX^\ast$ with the same utility.
Let $\va \in\cA_{\LDR}^{|\cK|}$ be the tuple specifying one action per type such that $\ve_{\va[k]}=\vy^k_{\vx}$ for each $k \in \cK$, \emph{i.e.}, each follower's type plays the best response for $\vx$.
Once we fix the best response for all the types, the objective is a linear function. Hence, it is linear on $\cX^{\va}$.
Then, there exists a vertex of $\cX^{\va}\subseteq \cX^*$ in which the objective is maximized.
Notice that in the vertex the follower could play a different best response. However, by the optimistic tie breaking assumption, with the best response the leader's utility increases, while the costs do not change.
This concludes the proof.
\end{proof}

\bigskip
\noindent{ \bf\  Lemma~\ref{lemma:stackelberg 2}.} 	
{\em It holds $|\cX^*| \le (|\cK|n_{\FLL}^2)^{n_{\LDR}-1}$.}
\begin{proof}{Proof.}
Each polytope $\cX^{\va}$, $\va \in \cA_{\LDR}^{|\cK|}$ is defined by the following inequalities over $\vx$ that imply the optimality of the tuple of best responses $\va$:
\[ \vx^\top U_{k} \ve_{\va[k]} \ge \vx^\top U_{k} \ve_{a} \ \ \forall k \in \cK, a \in \cA_{\FLL}.\]
Thus, each vertex $V(\cX^{\va})$, $\va \in \cA_{\LDR}^{|\cK|}$,  is the intersection of $n_{\LDR}-1$ equalities belonging to the following set:
\[ \vx^\top U_{k} \ve_{a}= \vx^\top U_{k} \ve_{a'} \ \ \forall k \in \cK,\, a,a' \in \cA_{\FLL},\]
and the simplex constraint.
Hence, there are at most $(|\cK|n_{\FLL}^2)^{n_{\LDR}-1}$ vertices.
\end{proof}

\noindent{ \bf\  Lemma~\ref{lemma:apxDiscretized}.} 	
{\em	For any sequence $(v_t,m_t,\lambda_t)_{t\in[T]}$ and $\epsilon\in[1/T,1]$, it holds
 	$$\sup_{\p^\ast\in\psl} \mleft\{\sum_{t \in [T]}\mleft( f_t(\p^\ast(v_t))- \lambda_t c_t(\p^\ast(v_t))\mright)\mright\}- \max_{\phi\in\Phi_\epsilon}\mleft\{ \sum_{t\in[T]}\mleft(f_t(\phi(v_t,\lambda_t))- \lambda_t c_t(\phi(v_t,\lambda_t))\mright)\mright\}\le \frac{7 \epsilon T}{\rho} .$$}

\begin{proof}{Proof.}
    Consider a policy $\p^\ast \in \psl$ such that \[\sum_{t \in [T]}\mleft( f_t(\p^\ast(v_t))- \lambda_t c_t(\p^\ast(v_t))\mright)\ge \sup_{\p \in\psl} \mleft\{\sum_{t \in [T]}\mleft( f_t(\p(v_t))- \lambda_t c_t(\p(v_t))\mright)\mright\}- \frac{\epsilon T}{\rho}.\]
 	Let $\hat\p$ be a policy such that $\hat\p(v)=\disc{\p^\ast(v)}_{1/T}$ for each $v\in\cV_{\epsilon}$. We have that $\hat\p\in\hat\ps_{\epsilon}$ since it is 2-Lipschitz-continuous: for each $v,v'\in\cV_\epsilon$, it holds 
 	\begin{align*}
 		\mleft|\hat\p(v) - \hat\pi(v') \mright| & = \mleft| \disc{\p^\ast(v)}_{1/T} - \disc{\p^\ast(v')}_{1/T}  \mright|\\
 		& \le \mleft| \p^\ast(v) - \p^\ast(v')  \mright| + \frac{1}{T}\\
 		&  \le \mleft|v-v'  \mright| + \frac{1}{T}\\
 		& \le 2 |v-v'|,
 	\end{align*}
 	where the first equality is by definition of $\hat\p$, the first inequality is by definition of discretization with step $1/T$ and by applying the triangle inequality, the second inequality is by 1-Lipschitz continuity of $\p^\ast$, and the last inequality holds because $|v-v'|\ge\epsilon$, and $\epsilon\in[1/T,1]$.
 	
 	Let $\phi:\cV\times\cD\to\cB$ be the thresholded  policy $\gamma(\hat\p)$. We show that, for each $t\in[T]$, the expected utility for the thresholded policy $\phi$ is \emph{close} to the expected utility guaranteed by $\p^\ast$, which is $\indicator{\p^*(v_t)\ge m_t} \cdot (v_t-(1+\lambda_t)\p^*(v_t))$. 
 	First, we observe that, by 1-Lipschitz continuity of $\pi^\ast$, we have that, for each $v\in\cV$, 
 	\[
 	\mleft| \p^\ast(\disc{v}_{\epsilon}) - \p^\ast(v) \mright| \le |\disc{v}_{\epsilon}-v|.
 	\] 
 	Then,
 	\[
 	-v+\disc{v}_{\epsilon} \le \p^\ast(\disc{v}_{\epsilon}) - \p^\ast(v) \le v-\disc{v}_{\epsilon}.
 	\]
 	The first inequality yields 
 	\begin{equation}\label{aux p ast disc1}
 		\disc{\p^\ast(\disc{v}_{\epsilon})}_{1/T}\ge \p^\ast(v) -\epsilon -\frac{1}{T}\ge \p^\ast(v)-2\epsilon.
 	\end{equation}

	The second inequality yields
 	\begin{equation}\label{aux p ast disc2}
 		\disc{\p^\ast(\disc{v}_{\epsilon})}_{1/T}\le  \p^\ast(\disc{v}_\epsilon) \le \p^\ast(v)  +\epsilon.
 	\end{equation}
 	Now, we distinguish two cases. 
 	
 	\textbf{Case $v_t\ge (1+\lambda_t)\p^\ast(v_t)$.} For each $t\in[T]$, by \cref{def:thresholded}, and by \cref{aux p ast disc1} we have
 	 \[
 	\phi(v_t,\lambda_t)=\min\mleft\{\disc{\p^\ast(\disc{v}_\epsilon)}_{1/T} + 2\epsilon \,,\, \frac{v_t}{1+\lambda_t} \mright\}\ge \p^\ast(v_t),
 	\]
 	and, by \cref{aux p ast disc2},
 	\[
 	\phi(v_t,\lambda_t)=\min\mleft\{\disc{\p^\ast(\disc{v}_\epsilon)}_{1/T} + 2\epsilon \,,\, \frac{v_t}{1+\lambda_t} \mright\}\le \p^\ast(v_t) + 3\epsilon.
 	\]
 	Then, by using the last two inequalities, the expected utility can be lower bounded as follows
 	\begin{align*}
 		\indicator{\phi(v_t,\lambda_t)\ge m_t}\mleft( v_t - (1+\lambda_t)\,\phi(v_t,\lambda_t) \mright) & \ge 	\indicator{\p^\ast(v_t)\ge m_t}\mleft( v_t - (1+\lambda_t)\,\phi(v_t,\lambda_t) \mright)\\
 		& \ge \indicator{\p^\ast(v_t)\ge m_t}\mleft( v_t - (1+\lambda_t)\,(\p^\ast(v_t)+3\epsilon) \mright) \\
 		& \ge \indicator{\p^\ast(v_t)\ge m_t}\mleft( v_t - (1+\lambda_t)\,\p^\ast(v_t) \mright) - 3\epsilon (1+\lambda_t).
 	\end{align*}
 
 \textbf{Case $v_t< (1+\lambda_t)\p^\ast(v_t)$.}	The expected utility of the primal player when employing policy $\p^\ast$ is less than or equal to $0$, while the expected utility for $\phi(v_t,\lambda_t)$ is at least $0$ since $(v_t-(1+\lambda_t)\,\phi(v_t,\lambda_t))\ge 0$ by construction (\cref{def:thresholded}).
 
Finally, by summing over all steps $t\in[T]$, we have
\begin{align*}
	\sum_{t \in [T]}\indicator{\phi(v_t,\lambda_t)\ge m_t} \, (v_t- (1+\lambda_t)\,\phi(v_t,\lambda_t)) & \ge  \sum_{t \in [T]}\mleft(\indicator{\p^\ast(v_t) \ge m_t} \, (v_t-(1+\lambda_t)\,\p^\ast(v_t)) -3\epsilon(1+\lambda_t)\mright)
	\\ &  \ge \sum_{t \in [T]}\mleft(\indicator{\pi^*(v_t)  \ge m_t} \, (v_t-(1+\lambda_t) \,\p^\ast(v_t))\mright) -3\epsilon T\mleft(1+\frac{1}{\rho}\mright) 
	\\ & \ge  \sum_{t \in [T]}\mleft(\indicator{\pi^*(v_t)  \ge m_t} \, (v_t-(1+\lambda_t) \,\p^\ast(v_t))\mright) -\frac{6\epsilon T}{\rho}.  
\end{align*}

The theorem follows by substituting the definition of $f_t$  and $c_t$ in the above expression and by the near-optimality of $\pi^\ast$. Formally,
\begin{align*}
\sum_{t\in[T]}\mleft(f_t(\phi(v_t,\lambda_t))- \lambda_t c_t(\phi(v_t,\lambda_t))\mright) &\ge \sum_{t \in [T]}\mleft( f_t(\p^\ast(v_t))- \lambda_t c_t(\p^\ast(v_t))\mright)-\frac{6\epsilon T}{\rho} \\
&\ge \sup_{\p \in\psl} \mleft\{\sum_{t \in [T]}\mleft( f_t(\p(v_t))- \lambda_t c_t(\p(v_t))\mright)\mright\}- \frac{7\epsilon T}{\rho}.
\end{align*}
This concludes the proof.

 \end{proof}

\noindent{ \bf\  Lemma~\ref{lemma:delta good}.}
{\em 	For each $m \in [M-1]$ and each $h \in \cH_m$, the edge $\zchild{h}$ is $\Delta_m$-good for $\Delta_m\defeq 4(1+1/\rho)\epsilon_m$.}
\begin{proof}{Proof.}
	Let $\pi'\in A(h)$ be an edge originating from node $h$ and $t \in T$. We need to prove that $ r(\zchild{h},\lambda_t,v_t,m_t) \ge  E_{\p \sim \vp_{t,h\cdot\pi'}}\mleft[r(\p,\lambda_t,v_t,m_t)\mright]  -\Delta_{m}$.
	Fist, we notice that it is sufficient to prove that \[r(\zchild{h},\lambda_t,v_t,m_t) \ge r(\bar \p,\lambda_t,v_t,m_t) -\Delta_{m}\] for each $\bar \pi$ in the support of $\vp_{t,h\cdot\pi'}$.
	By the structure of the tree we have that $\zchild{h}(\lfloor v_t\rfloor_{\epsilon_m})=\bar \p(\lfloor v_t\rfloor_{\epsilon_m})$. Hence, 
	\begin{align*}
	    \zchild{h}(\lfloor v_t\rfloor_{\epsilon_M})&=\zchild{h}(\lfloor v_t\rfloor_{\epsilon_m})+ \disc{2(\disc{v_t}_{\epsilon_M}-\disc{v_t}_{\epsilon_m})}_{1/T} \\
	    &= \bar \p(\lfloor v_t\rfloor_{\epsilon_m})+ \disc{2(\disc{v_t}_{\epsilon_M}-\disc{v_t}_{\epsilon_m})}_{1/T} \\
	    & = \bar \p(\lfloor v_t\rfloor_{\epsilon_M})- (\bar \p(\lfloor v_t\rfloor_{\epsilon_M})+\bar \p(\lfloor v_t\rfloor_{\epsilon_m}) )+ \disc{2(v_t-\disc{v_t}_{\epsilon_m})}_{1/T}\\
	    & = \bar \p(\lfloor v_t\rfloor_{\epsilon_M})- \disc{ (\bar \p(\lfloor v_t\rfloor_{\epsilon_M})+\bar \p(\lfloor v_t\rfloor_{\epsilon_m}) )}_{1/T}+ \disc{2(\disc{v_t}_{\epsilon_M}-\disc{v_t}_{\epsilon_m})}_{1/T}\\
	    & \ge \bar \p(\lfloor v_t\rfloor_{\epsilon_M}) - \disc{2(\disc{v_t}_{\epsilon_M}-\disc{v_t}_{\epsilon_m})}_{1/T}+\disc{2(\disc{v_t}_{\epsilon_M}-\disc{v_t}_{\epsilon_m})}_{1/T}\\
	    & = \bar \p(\lfloor v_t\rfloor_{\epsilon_M}),
	\end{align*}
    where the second-to-last equality holds since  $\p(\lfloor v_t\rfloor_{\epsilon_M}),\bar \p(\lfloor v_t\rfloor_{\epsilon_m})\in \cB_{1/T} $.
	Moreover, 
	\begin{align*}
	    \zchild{h}(\lfloor v_t\rfloor_{\epsilon_M})&=\zchild{h}(\lfloor v_t\rfloor_{\epsilon_m})+ \disc{2(\disc{v_t}_{\epsilon_M}-\disc{v_t}_{\epsilon_m})}_{1/T} \\
    	&= \bar \p(\lfloor v_t\rfloor_{\epsilon_m})+ \disc{2(\disc{v_t}_{\epsilon_M}-\disc{v_t}_{\epsilon_m})}_{1/T} \\
	    & = \bar \p(\lfloor v_t\rfloor_{\epsilon_M})- (\bar \p(\lfloor v_t\rfloor_{\epsilon_M})+\bar \p(\lfloor v_t\rfloor_{\epsilon_m}) )+ \disc{2(\disc{v_t}_{\epsilon_M}-\disc{v_t}_{\epsilon_m})}_{1/T}\\
	    & \le  \bar \p(\lfloor v_t\rfloor_{\epsilon_M}) +2\epsilon_m+\disc{2(\disc{v_t}_{\epsilon_M}-\disc{v_t}_{\epsilon_m})}_{1/T}\\
	    & \le  \bar \p(\lfloor v_t\rfloor_{\epsilon_M}) + 4 \epsilon_m.
	\end{align*}
    
    Let $\phi_h=\gamma(\zchild{h})$ and $\bar \phi= \gamma(\bar \p)$.
    We have that
    \[\phi_h(v_t,\lambda_t)= \min\mleft\{(\zchild{h}(\lfloor v_t\rfloor_{\epsilon_M}) + 2\epsilon_M\,,\,\frac{v_t}{1+\lambda_t}\mright\}\ge  \min\mleft\{(\bar \p(\lfloor v_t\rfloor_{\epsilon_M}) + 2\epsilon_M\,,\,\frac{v_t}{1+\lambda_t}\mright\} = \bar \phi(v_t,\lambda_t),\]
    and 
    \begin{align*}
        \phi_h(v_t,\lambda_t)&= \min\mleft\{(\zchild{h}(\lfloor v_t\rfloor_{\epsilon_M}) + 2\epsilon_M\,,\,\frac{v_t}{1+\lambda_t}\mright\}\\
        & \le \min\mleft\{(\bar \p(\lfloor v_t\rfloor_{\epsilon_M})+ 4\epsilon_m + 2\epsilon_M\,,\,\frac{v_t}{1+\lambda_t}\mright\} \\
        & \le \min\mleft\{(\bar \p(\lfloor v_t\rfloor_{\epsilon_M}) + 2\epsilon_M\,,\,\frac{v_t}{1+\lambda_t}\mright\} + 4\epsilon_m\\
        &= \bar \phi(v_t,\lambda_t) + 4\epsilon_m.
    \end{align*}
    
    Finally,
    \begin{align*}
 	r(\zchild{h},\lambda_t,v_t,m_t)&=	\indicator{\phi_h(v_t,\lambda_t)\ge m_t}\mleft( v_t - (1+\lambda_t)\,\phi_h(v_t,\lambda_t) \mright) \\
 	& \ge 	\indicator{ \bar \phi(v_t,\lambda_t)\ge m_t}\mleft( v_t - (1+\lambda_t)\, \phi_h(v_t,\lambda_t) \mright)\\
 	& \ge \indicator{\bar \phi(v_t,\lambda_t)\ge m_t}\mleft( v_t - (1+\lambda_t)\,(\bar \phi(v_t)+4\epsilon) \mright) \\
 	& \ge \indicator{\bar \phi(v_t,\lambda_t)\ge m_t}\mleft( v_t - (1+\lambda_t)\,(\bar \phi(v_t) \mright) - 4\epsilon (1+\lambda_t)\\
 	& \ge\indicator{\bar \phi(v_t,\lambda_t)\ge m_t}\mleft( v_t - (1+\lambda_t)\,(\bar \phi(v_t) \mright) - 4\epsilon \mleft(1+\frac{1}{\rho}\mright)\\
 	&=r(\bar \p,\lambda_t,v_t,m_t)-4\epsilon \mleft(1+\frac{1}{\rho}\mright)\\
 	&=r(\bar \p,\lambda_t,v_t,m_t)-\Delta_m,
 	\end{align*}
 	where the first inequality holds since $\phi_h(v_t,\lambda_t)\ge \bar \phi(v_t,\lambda_t)$, the second since $\phi_h(v_t,\lambda_t)\le \bar \phi(v_t,\lambda_t) +4\epsilon$, and the fourth since $\lambda_t \le 1/\rho$.
 	This concludes the proof.
\end{proof}

\noindent{ \bf\  Lemma~\ref{lemma:regret good expert}.}
{\em Given $m \in [M]$ and a node $h \in \mathcal{H}_m$, the exponential-weights policy with a time-varying learning rate $\eta_t = \min\{1/4, \sqrt{\log |A(h)|/(t\Delta_m)}\}$ guarantees the following regret upper bound: 
	\[
	R_{T}^h \le 4\sqrt{T\Delta_m\log |A(h)|)} + 32(4+\log T)\log |A(h)|. 
	\]
}
\proof{Proof.}
    The lemma follows directly from Theorem~3 in \cite{han2020learning}. In particular, it is sufficient to observe that the regret minimizer at node $h$ has to choose among $|A(h)|$ experts, and that there exists a $\Delta_m$-good expert by \cref{lemma:delta good}. Moreover, the payoff of each expert is in $[0,1]$ since all the considered policies are thresholded.
\endproof

\noindent{ \bf\  Theorem~\ref{thm:firstPrice}.}
{\em Algorithm~\ref{alg:first price} guarantees regret
	    \begin{align*}
	        \regp[T]\le  O\mleft( \frac{1}{\rho}\sqrt{T} \log^2(T)\mright).
	    \end{align*}  
}

\begin{proof}{Proof.}
	Let $\phi^\ast \in \Phi_{\epsilon_M}$ be an optimal policy in $\Phi_{\epsilon_M}$, that is 
	\[\phi^* \in \argmax_{\phi\in\Phi_{\epsilon_M}}\mleft\{ \sum_{t\in[T]}\mleft(f_t(\phi(v_t,\lambda_t))- \lambda_t c_t(\phi(v_t,\lambda_t))\mright)\mright\}.\]
	Let $\pi^\ast \in \hat \ps_{\epsilon_M}$ be a discrete policy such that $\gamma(\pi^\ast)=\phi^*$. This policy is guaranteed to exist by construction. Moreover, there exists a terminal node $h_M\in\cZ$ (i.e., a node at depth $M$) such that $\sigma_{h_M}=\pi^\ast$. Finally, let $\hroot,h_1, \dots, h_M$ be the sequence of nodes on the path going from the root $\hroot$ to the terminal node $h_M$.
	
	For each $t\in [T]$ we have that
	\[
	r_t(\pi^\ast)-\E_{\pi\sim \vp_{t,\hroot}}\mleft[r_t(\pi)\mright]=\sum_{m\in[M]}\mleft( \E_{\pi \sim \vp_{t,h_{m}}}\mleft[r_t(\pi)\mright] - \E_{\pi \sim \vp_{t,h_{m-1}}}\mleft[r_t(\pi)\mright] \mright),
	\]
	where we recall that the only policy in the support of $\vp_{t,h_{M}}$ is $\pi^\ast$.
	This implies
	\begin{align} 
		\sum_{t \in [T]}\mleft( r_t(\pi^\ast) - \E_{ \pi \sim \vp_{t,\hroot}}\mleft[r_t(\pi)\mright]\mright) & =  \sum_{t \in [T]} \sum_{m \in [M]} \mleft( \E_{\pi \sim \vp_{t,h_{m}}}\mleft[r_t(\pi)\mright] - \E_{\pi \sim \vp_{t,h_{m-1}}}\mleft[r_t(\pi)\mright] \mright)\nonumber\\
		& = \sum_{m \in [M]}  \sum_{t \in [T]} \mleft( \E_{\pi \sim  \vp_{t,h_{m}}}\mleft[r_t(\pi)\mright] - \E_{\pi \sim  \vp_{t,h_{m-1}}}\mleft[r_t(\pi)\mright] \mright).\label{eq:regretFirst}
	\end{align}
    Then, in order to upper bound the cumulative regret of the primal regret minimizer $\regp[T]$, we need to provide an upper bound to
    $
    \sum_{m \in [M]}  \sum_{t \in [T]} \mleft( \E_{\pi \sim \vp_{t,h_{m}}}\mleft[r_t(\pi)\mright] - \E_{ \pi \sim \vp_{t,h_{m-1}}}\mleft[r_t(\pi)\mright] \mright).
    $
    Given $m\in[M]$, we have 
    \begin{align}
    \sum_{t \in [T]} \mleft( \E_{\pi \sim \vp_{t,h_{m}}}\mleft[r_t(\pi)\mright] - \E_{ \pi \sim  \vp_{t,h_{m-1}}}\mleft[r_t(\pi)\mright] \mright)
    &=\sum_{t \in [T]} \mleft( \E_{\pi \sim  \vp_{t,h_{m}}}\mleft[r_t(\pi)\mright] - \E_{\pi\sim \vq_{t,h_{m-1}}}\E_{ \pi'\sim \vp_{t,h_{m-1}\cdot\pi}}\mleft[r_t(\pi')\mright] \mright)\nonumber\\
    &\le  4\sqrt{T\Delta_{\epsilon_{m-1}}\log |A(h_{m-1})|} + 32(4+\log T)\log |A(h_{m-1})|, \label{eq:regretSecond}
    \end{align}
    where the inequality follows from Theorem~\ref{lemma:regret good expert}.

    Now, we can derive the regret upper bound. By definition of cumulative regret of the primal player,
    \[
        \regp[T] = \sup_{\p\in\psl} \mleft\{\sum_{t \in [T]}\mleft( f_t(\p(v_t))- \lambda_t c_t(\p(v_t))\mright)\mright\} - \sum_{t\in[T]} \E_{\p \sim \vp_{t,\hroot}}\mleft[r_t(\pi)\mright].
    \]
    Then, by Lemma~\ref{lemma:apxDiscretized} and \cref{eq:regretFirst} we have 
    \begin{align*}
    \regp[T] & \le \frac{7 \epsilon_M T}{\rho} + \sum_{t \in [T]}r_t(\pi^\ast)- \sum_{t\in[T]} \E_{\p \sim \vp_{t,\hroot}}\mleft[r_t(\pi)\mright]\\ & =  \underbrace{\frac{7 \epsilon_M T}{\rho} +  \sum_{m \in [M]}  \sum_{t \in [T]} \mleft( \E_{\pi \sim \vp_{t,h_{m}}}\mleft[r_t(\pi)\mright] - \E_{\pi \sim \vp_{t,h_{m-1}}}\mleft[r_t(\pi)\mright] \mright)}_{\circled{C}}.
    \end{align*}

    Then, we can bound $\circled{C}$ as follows
    \begin{subequations}
    \begin{align}
               \circled{C}& \le \frac{7 \epsilon_M T}{\rho} + \sum_{m \in [M]}\mleft( 4\sqrt{T\Delta_{\epsilon_{m-1}}\log |A(h_{m-1})|} + 32(4+\log T)\log |A(h_{m-1})|\mright) \label{eq:final3}\\
               & \le \frac{7 \epsilon_M T}{\rho} + \sum_{m \in [M]}\mleft( 4\sqrt{T\Delta_{\epsilon_{m-1}}\log T^{\frac{2}{\epsilon_{m}}}} + 32(4+\log T)\log T^{\frac{2}{\epsilon_{m}}}\mright) \label{eq:final4}\\
               &\le \frac{7 \epsilon_M T}{\rho} + \sum_{m \in [M]}\mleft( 4\sqrt{T 4 (1+1/\rho)\epsilon_{m-1} \log T^{\frac{2}{\epsilon_{m}}}} + 32(4+\log T)\log T^{\frac{2}{\epsilon_{m}}}\mright) \label{eq:final5}\\
               & = \frac{7 \epsilon_M T}{\rho} + \sum_{m \in [M]}\mleft( 4\sqrt{16 T  (1+1/\rho) \log T} + 64(4+\log T) \frac{1}{\epsilon_m}\log T\mright)\label{eq:final6}\\
               & = \frac{7 \epsilon_M T}{\rho} +   2\log(T)\sqrt{16 T  (1+1/\rho) \log T} + 64(4+\log T) \log T \sum_{m \in [M]}\frac{1}{\epsilon_m}\label{eq:final7}\\ 
               & \le \frac{7 \epsilon_M T}{\rho} +   2\log(T)\sqrt{16 T  (1+1/\rho) \log T} + 64(4+\log T) \log T \frac{1}{\epsilon_{M+1}}\label{eq:final8}\\
               &=\frac{7 \sqrt{ T}}{\rho} +   2\log(T)\sqrt{16 T  (1+1/\rho) \log T} + 128(4+\log T)\log T  \sqrt{T} \nonumber\\
               &\le O\mleft( \frac{\sqrt{T} \log^2(T)}{\rho}\mright),\nonumber
    \end{align}
    \end{subequations}

    where \cref{eq:final3} follows from \cref{eq:regretSecond}, \cref{eq:final4} follows from $|A(h_{m-1})|\le T^{\frac{2}{\epsilon_{m}}}$, \cref{eq:final5} follows from the definition of $\Delta_m$, \cref{eq:final6} follows from $\epsilon_m=2 \epsilon_{m+1}$, \cref{eq:final7} follows from the definition of $M$, and \cref{eq:final8} follows from $\sum_{m \in [M]}\frac{1}{\epsilon_m} \le \frac{1}{\epsilon_{M+1}}$.
    This concludes the proof.
\end{proof}

\section{On Strong Duality in Semi-Infinite LPs}\label{appendix strong duality}

We provide a simple example in which a semi-infinite linear optimization problem does not admit strong duality. 
Let $\cX=[0,1]$. Define $f:\cX\to\R$ by
\[
    f(x)\defeq\mleft\{\hspace{-1.25mm}\begin{array}{l}
        \displaystyle
        1 \hspace{.5cm}\text{\normalfont if } x=0\\ [2mm]
        0 \hspace{.5cm}\text{\normalfont if } x\in(0,1]
    \end{array}\mright..
\]
Then, let $\Xi=\Delta^\cX$ and consider the linear program
\begin{equation}
    \mleft\{\hspace{-1.25mm}\begin{array}{l}
        \displaystyle
        \inf_{\xi\in\Xi}\E_{x\sim\xi}\mleft[f(x)\mright] \\ [2mm]
        \displaystyle \text{\normalfont s.t. } \E_{x\sim\xi}\mleft[x\mright]\le 0
    \end{array}\mright..
\end{equation}
Since $x\in[0,1]$, the only way in which the constraint can be satisfied is always selecting $x=0$. Then, the primal optimal value is $p^\ast=1$. Now, the Lagrangian dual of the problem is
$
g(\lambda)=\inf_{\xi\in\Xi}\mleft\{ \E\mleft[f(x)\mright] + \lambda\E\mleft[x\mright]\mright\}.
$
We have $d^\ast=\sup_{\lambda\ge 0} g(\lambda)=0$. Therefore, we have a duality gap of 1.